\documentclass[journal]{IEEEtran}
\usepackage{color}
\usepackage{graphicx}
\usepackage{subfigure}
\usepackage{threeparttable}
\usepackage{latexsym}
\usepackage{amsfonts}
\usepackage{amsmath}
\usepackage{multirow}
\usepackage{multicol}
\usepackage{makecell}
\usepackage{booktabs}
\usepackage{graphicx}
\usepackage{microtype}
\usepackage{url}

\newcommand{\tabincell}[2]{\begin{tabular}{@{}#1@{}}#2\end{tabular}}
\usepackage{algorithm}
\usepackage{algorithmic}

\newcommand{\bm}[1]{\mbox{\boldmath{$#1$}}}

\usepackage{cite}

\ifCLASSINFOpdf
\else
\fi
\begin{document}
%
\title{Learning to Classify Open Intent via Soft Labeling and Manifold Mixup}
%
%
%

\author{Zifeng Cheng, Zhiwei Jiang, Yafeng Yin, Cong Wang, and Qing Gu
\thanks{Manuscript received August 3, 2021; revised November 28, 2021; accepted January 6, 2022.
This work was supported by the National Science Foundation of China under Grants 61906085, 62172208, 61972192, 61802169, 41972111;
the Second Tibetan Plateau Scientific Expedition and Research Program under Grant 2019QZKK0204.
This work is partially supported by Collaborative Innovation Center of Novel Software Technology and Industrialization. (Corresponding author: Zhiwei Jiang.)}
\thanks{The authors are with the State Key Laboratory for Novel Software Technology, Nanjing University, Nanjing 210023, China (e-mail: chengzf@smail.nju.edu.cn; jzw@nju.edu.cn; yafeng@nju.edu.cn; cw@smail.nju.edu.cn; guq@nju.edu.cn).}
}

%
%

\markboth{IEEE/ACM TRANSACTIONS ON AUDIO, SPEECH, AND LANGUAGE PROCESSING, VOL. XX, 2021}%
{Shell \MakeLowercase{\textit{et al.}}: Bare Demo of IEEEtran.cls for IEEE Journals}
%



\maketitle


\begin{abstract}
Open intent classification is a practical yet challenging task in dialogue systems.
Its objective is to accurately classify samples of known intents while at the same time detecting those of open (unknown) intents.
Existing methods usually use outlier detection algorithms combined with K-class classifier to detect open intents, where K represents the class number of known intents.
Different from them, in this paper, we consider another way without using outlier detection algorithms.
Specifically, we directly train a (K+1)-class classifier for open intent classification, where the (K+1)-th class represents open intents.
To address the challenge that training a (K+1)-class classifier with training samples of only K classes, we propose a deep model based on Soft Labeling and Manifold Mixup (SLMM).
In our method, soft labeling is used to reshape the label distribution of the known intent samples, aiming at reducing model's overconfident on known intents.
Manifold mixup is used to generate pseudo samples for open intents, aiming at well optimizing the decision boundary of open intents.
Experiments on four benchmark datasets demonstrate that our method outperforms previous methods and achieves state-of-the-art performance.
All the code and data of this work can be obtained at \url{https://github.com/zifengcheng/SLMM}.
\end{abstract}

\begin{IEEEkeywords}
Open intent classification, soft labeling, manifold mixup
\end{IEEEkeywords}

%
\IEEEpeerreviewmaketitle

%
%
%
%

\section{Introduction}
\begin{figure}[t]
\centering
\includegraphics[width=0.8\columnwidth]{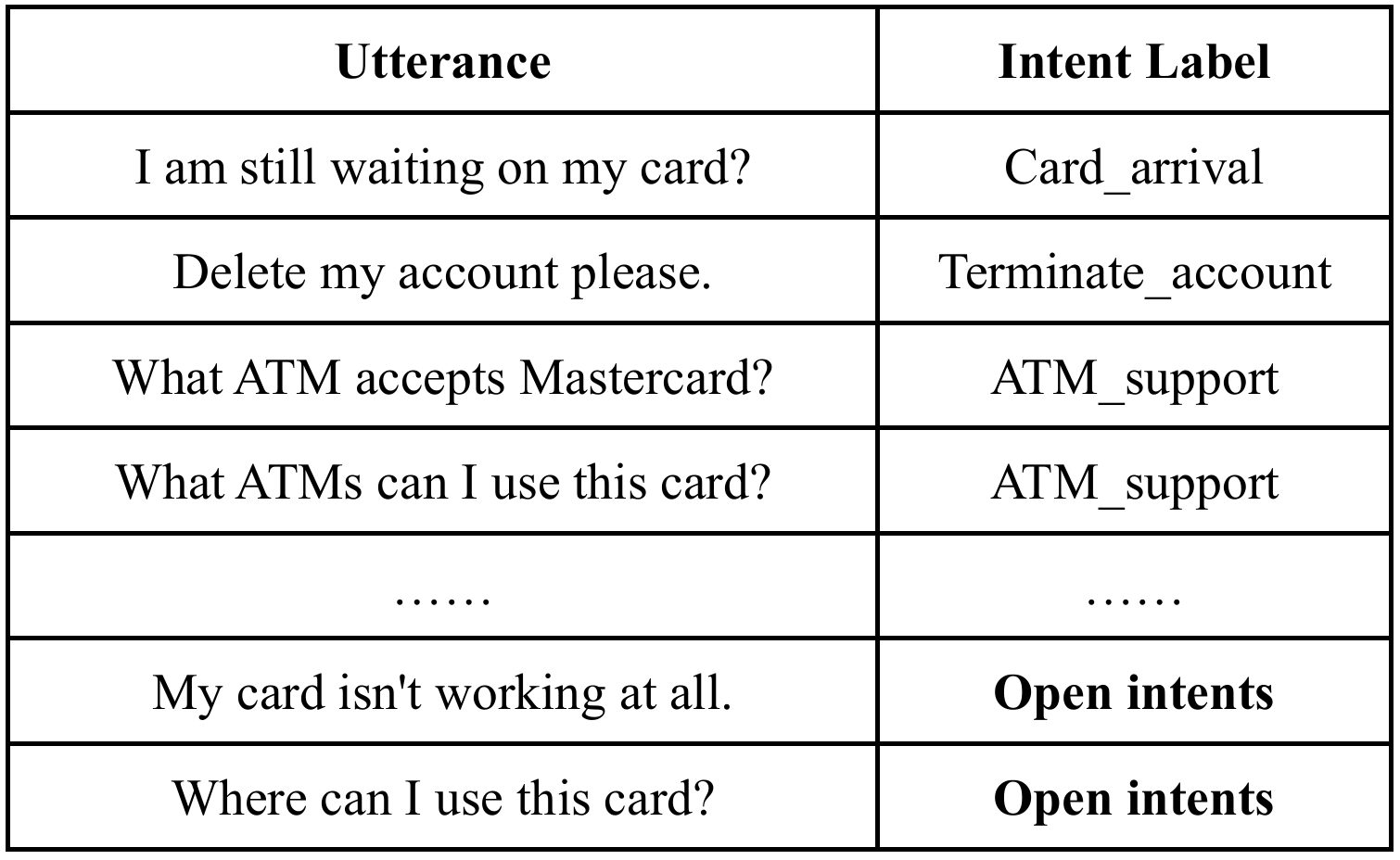}
\caption{An example of open intent classification in banking domain. Card\_arrival, Terminate\_account, and ATM\_support are three known intents. Other unsupported intents are treated as open intents.}  \label{fig:example}
\end{figure}

\IEEEPARstart{A}{ccurately} identifying user intents from utterances plays a critical role in task-oriented dialogue systems.
Recent years have witnessed the rapid validation of intent detection models \cite{DBLP:conf/emnlp/LiLQ18,DBLP:conf/acl/ENCS19,DBLP:conf/emnlp/QinCLWL19,DBLP:conf/acl/ZhangLDFY19,Qin21}.
While most of these models identify user intents via conducting multi-class classification on known intents, they cannot reject unsupported open (unknown) intents (i.e., they work with $closed$-$world$ assumption that the classes appeared in the test data must have appeared in training).
However, in practice, it is impossible to cover all user intents during training phase.
Therefore, to avoid performing wrong actions in response to user intents, it is crucial to effectively detect open intents.
Taking the dialogue system of banking domain as an example, as shown in Figure~\ref{fig:example}, the training set may only contain three supported known intents (i.e., Card\_arrival, Terminate\_account, and ATM\_support).
But in the testing phase, the model should not only correctly classify these three known intents, but also detect open intents unseen in the training set.
To this end, the task of open intent classification was proposed and has attracted a lot of attention recently \cite{Lin19,Yan20,DBLP:conf/emnlp/CavalinRAP20,DBLP:conf/coling/XuHYLLX20,zhang21,DBLP:conf/naacl/ZengHYXX21,Tan19,DBLP:conf/emnlp/LarsonMPCLHKLLT19,DBLP:journals/taslp/ZhengCH20,Zhang20}, which aims to conduct classification on samples of known intents while at the same time detecting those of open intents.

\begin{figure}[t]
\centering
\includegraphics[width=1\columnwidth]{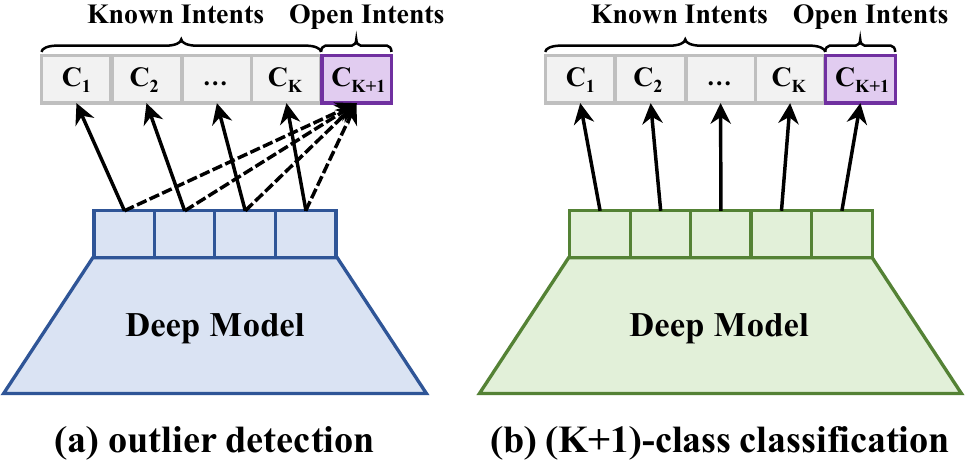}
\caption{Illustration of two kinds of deep open intent classification models. The left part is the idea of existing methods and the right part is our method.}
\label{fig:motivation}
\end{figure}

Generally, the task of open intent classification can be viewed as a (K+1)-class classification problem, where the first K classes represent K known intents and the (K+1)-th class represents open intents.
For this task, the main challenge is how to detect open intents without any training samples of open intents.
To address this challenge, recent studies mainly focused on designing outlier detection algorithms combined with K-class classifier.
In this kind of methods, as shown in the left part of Figure \ref{fig:motivation}, the open intents can be detected by judging whether the input sample is an outlier of all known intents.

Along this line of work, researchers mainly consider how to calibrate the decision boundary of each known intent class for outlier detection.
An intuitive solution is to simply use a threshold on the K-class classifier's prediction probability to decide whether a sample is an outlier of all known intents \cite{Shu17,DBLP:conf/emnlp/CavalinRAP20}.
However, since the deep learning methods tend to overfit the training samples, the K-class classifier would produce overconfident predictions of known intents \cite{DBLP:conf/iclr/LiangLS18,DBLP:conf/cvpr/Hein}.
Thus, even the sample of open intents would be easily classified with high probability into the K known intents, which makes the threshold hard to decide.
To this end, some subsequent work chooses to use more flexible outlier detection algorithms to calibrate the decision boundary.
For example, Lin et al. \cite{Lin19} and Yan et al. \cite{Yan20} propose to first learn discriminative deep features through large margin cosine loss and Gaussian mixture loss, and then apply the local outlier factor algorithm to detect open intents.
Xu et al. \cite{DBLP:conf/coling/XuHYLLX20} also proposes to first learn discriminative deep features through large margin cosine loss, and then apply Mahalanobis distance to detect open intents.
Considering the feature learning is optimized separately from decision boundary learning in these methods, Zhang et al. \cite{zhang21} proposes a joint optimized adaptive decision boundary method for outlier detection.

In this paper, we consider another way without using outlier detection algorithms.
As shown in the right part of Figure \ref{fig:motivation}, instead of combining K-class classifier with outlier detection algorithms, we directly train a (K+1)-class classifier for open intent classification.
Training a (K+1)-class classifier with training samples of only K classes is very challenging.
Obviously, there are two main challenges: one is how to calibrate the decision boundary of the K known intent classes to avoid the overconfident prediction on known intents, and the other is how to learn the decision boundary of the (K+1)-th open intent class to enable test samples to be classified as open intents.
For the first challenge, our intuition is whether we can reshape the label distribution of training samples of known intents to reduce overconfident prediction on known intents.
For the second challenge, our intuition is whether we can generate some pseudo samples for open intents based on existing training samples to optimize the decision boundary of open intents.

To this end, we propose a deep open intent classification model based on Soft Labeling and Manifold Mixup (SLMM).
Specifically, for the overconfident prediction problem, we design a Soft Labeling (SL) strategy to reshape the label distribution of training samples, which reallocates the probability of the known intent samples on the class of known intents to the class of open intents.
Soft labeling allows the model to give each sample a probability of being predicted as an open intent, thus reducing overconfident prediction on known intents.
For the optimization problem of open intents, we design a Manifold Mixup (MM) strategy to generate pseudo samples for open intents, which interpolates the representation of two samples of different known intents.
While the interpolation between two different known intents can be regarded as a low-confidence area of both known intents, we use the samples in these low-confidence areas as pseudo open intent samples to learn the decision boundary of open intents.
Based on these two strategies, our model can be trained with label-reshaped samples of known intents and pseudo samples of open intents, and thus can be effectively used for open intent classification.

The major contributions of this paper are summarized as follows:
\begin{itemize}
\item We propose a (K+1)-class classification framework for the task of open intent classification without using outlier detection algorithms.
\item We propose two strategies (i.e., soft labeling and manifold mixup) to learn decision boundary without using any additional data.
\item We conduct experiments on four benchmark datasets and the experimental results demonstrate that our model SLMM outperforms previous methods and achieves state-of-the-art performance.
\end{itemize}

\begin{figure*}[t]
\centering
\includegraphics[width=2\columnwidth]{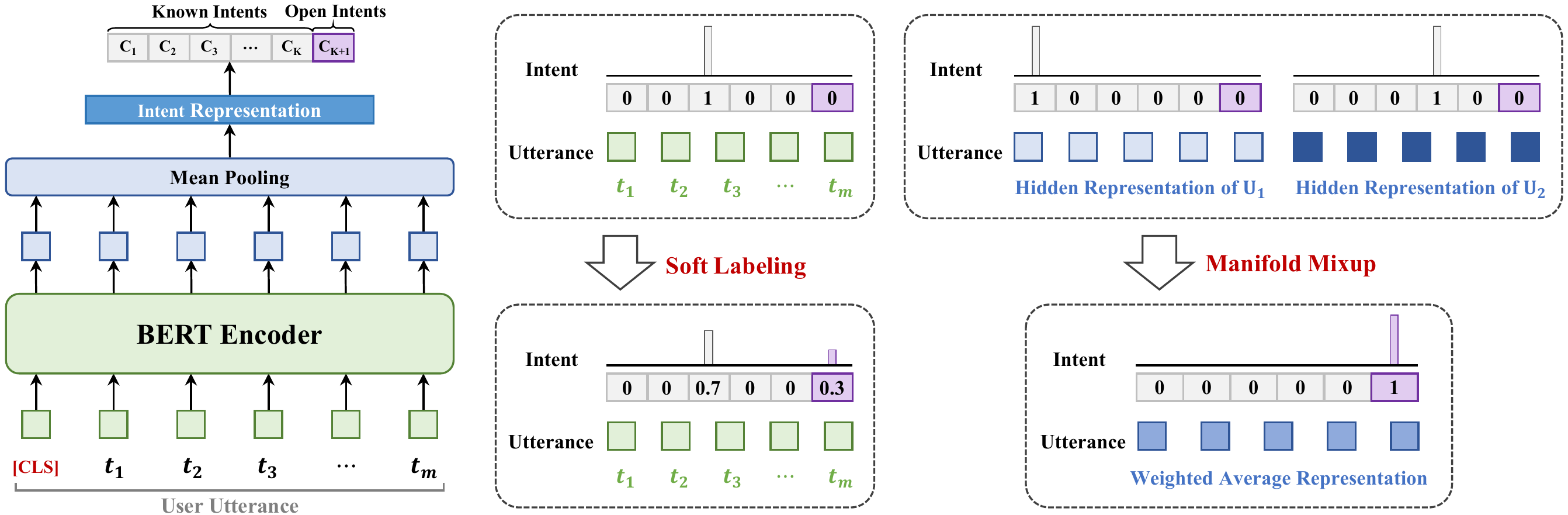}
\caption{Framework of our proposed method. The left part is model backbone in our proposed method. The middle part is soft labeling strategy and the right part is manifold mixup strategy.}
\label{fig:model}
\end{figure*}

\section{Related Work}
In this section, we introduce the following two research topics relevant to our work.

\subsection{Open Intent Classification}
Intent detection is an important component of dialogue system.
Many methods have been proposed to solve this task in recent years \cite{DBLP:conf/emnlp/LiLQ18,DBLP:conf/acl/ENCS19,DBLP:conf/emnlp/QinCLWL19,DBLP:conf/acl/ZhangLDFY19,Qin21} and most of these methods work well with the $closed$-$world$ assumption.
However, such an assumption is commonly violated in practical systems that are deployed in a dynamic or open environment.
Practical systems often encounter queries that fall outside their supported intents.
Therefore, recognizing queries that fall outside supported intents has gained attention recently.

In the past few years, researchers have proposed various methods to deal with this open intent detection problem.
The first group of methods mainly focuses on the out-of-domain intent detection \cite{DBLP:conf/emnlp/RyuKYL18,Kim18,DBLP:conf/aaai/PodolskiyLBAP21,Xu21}.
This task can be formalized as a binary classification problem, which aims to distinguish between in-domain data and out-of-domain data and does not require the more fine-grained classification of known intents.

Unlike these methods, the second group of methods expects to be able to detect the out-of-domain intents while at the same time performing more fine-grained classification for known intents.
In these methods, some labeled open intents samples are available in training data~\cite{DBLP:conf/emnlp/LarsonMPCLHKLLT19,DBLP:journals/taslp/ZhengCH20,DBLP:conf/naacl/ZengHYXX21}.
However, under open environment, collecting data of open intents is very difficult and expensive.
Thus, the third group of methods is proposed to focus on a more practical setting, which expects to train the open intent classification model without using any labeled samples of open intents in training data~\cite{Lin19,Yan20,DBLP:conf/emnlp/CavalinRAP20,DBLP:conf/coling/XuHYLLX20,zhang21,Tan19,Zhang20}.
Considering the scarcity of data in practice, Tan et al.~\cite{Tan19} and Zhang et al.~\cite{Zhang20} explore the few-shot scenario (i.e., the samples of known intents are few-shot).

Among the third group of methods, current methods mainly employ the outlier detection algorithms for open intent detection.
These methods can be further divided into three categories: threshold-based, post-processing, and joint-optimization.
The threshold-based methods simply use a threshold of prediction probability to judge whether the example belongs to open intents.
Cavalin et al. \cite{DBLP:conf/emnlp/CavalinRAP20} designs such a method, which is based on a special word graph and the threshold is applied to the maximum prediction probability of nodes in graph.
The post-processing methods first learn a feature space based on training samples of known intents and then apply the outlier detection algorithm on the feature space to detect open intents in a post-processing way.
Lin et al. \cite{Lin19} proposes a two-step method which uses large margin cosine loss (LMLC) to learn discriminative features and then uses a density-based detection algorithm LOF to detect open intents.
Yan et al. \cite{Yan20} proposes a semantic-enhanced Gaussian mixture model to learn discriminative features and also uses the LOF algorithm to detect open intents.
Xu et al. \cite{DBLP:conf/coling/XuHYLLX20} proposes BiLSTM with LMLC as a feature extractor and uses Gaussian discriminant analysis (GDA) and Mahalanobis distance to detect open intents.
Similar to the post-processing methods, the joint-optimization methods also detect open intents in a post-processing way, but they consider to jointly learn the feature space and the decision boundary of outlier detection.
Zhang et al. \cite{zhang21} proposes a joint optimized adaptive decision boundary method for open intent classification.

\subsection{Mixup}
Mixup \cite{ZhangCDL18} is a recent proposed data augmentation method, which can construct virtual samples by linear interpolation between two random samples.
As a special mixup method, manifold mixup \cite{DBLP:conf/icml/VermaLBNMLB19} is also able to construct virtual samples, but its specialness is that it constructs samples by interpolating the hidden representation of samples.
Since manifold mixup often leads to smoother decision boundaries that are further away from the training data, it is usually used to improve the generalization of neural network.
While these mixup methods are mainly adopted in the field of computer vision, recently, some work has tried to use them in the field of natural language processing \cite{DBLP:conf/acl/ChengJME20,DBLP:conf/acl/ChenYY20,DBLP:conf/emnlp/ZhangYZ20,DBLP:conf/emnlp/ChenWTYY20,DBLP:conf/www/MiaoLWT20}.
Among them, Cheng et al. \cite{DBLP:conf/acl/ChengJME20} performs interpolations at the embedding space in sequence-to-sequence learning for machine translation.
Chen et al. \cite{DBLP:conf/acl/ChenYY20} and Miao et al. \cite{DBLP:conf/www/MiaoLWT20} explore mixup on semi-supervised text classification task.
Chen et al. \cite{DBLP:conf/emnlp/ChenWTYY20} and Zhang et al. \cite{DBLP:conf/emnlp/ZhangYZ20} explore interpolation on sequence labeling tasks.
Different from most previous studies that use mixup methods to augment data and improve model generalization, we adopt manifold mixup in the task of open intent classification and mainly use it to generate samples for previous unseen intents.
In the same period of our work, some work \cite{DBLP:conf/cvpr/ZhouYZ21,DBLP:conf/acl/ZhanLLFWL21} use mixup to generate data for open class and model this task as (K+1)-class classification.
Zhou et al. \cite{DBLP:conf/cvpr/ZhouYZ21} learns placeholders for open-set recognition, consisting of classifier placeholder and data placeholder.
They use adaptive threshold to learn classifier placeholder and use mixup to learn data placeholder.
Zhan et al. \cite{DBLP:conf/acl/ZhanLLFWL21} constructs two types of pseudo outliers by using mixup and leveraging publicly available auxiliary datasets to form a consistent (K+1)-class classification task.

\section{Method}
In this section, we first present the task definition of open intent classification.
Then we introduce the overview of our proposed method and the detailed model architecture.
Finally, we detail the training and inference process of the method. 
\subsection{Task Definition}
We first introduce some notation and formalize the open intent classification task.
Let $\mathcal{D}_{tr} = \{(x_i,y_i)\}_{i=1}^{L}$ denote the training set, where $x_i$ is a training utterance (sample), $y_i \in Y = \{1,2,\cdots,K\}$ is the ground-truth intent (label) of $x_i$, and $L$ is the number of utterances in the training set.
Let $\mathcal{D}_{te} = \{(x_j,y_j)\}_{j=1}^{U}$ denote a test set, where $x_j$ is a test utterance, $y_j \in \hat{Y} = \{1,2,\cdots,K,K+1\}$ is the ground-truth intent of $x_j$, and $U$ is the number of utterances in the test set.
Note that the class $K+1$ in $\hat{Y}$ is a proxy class for open intents which has not been seen in the training set $\mathcal{D}_{tr}$.
Then, the goal of open intent classification is to learn a model based on the training set $\mathcal{D}_{tr}$ and apply it on the test set $\mathcal{D}_{te}$ to (1) predict the correct intents for utterances of known intents, and (2) identify utterances of open intents.

\subsection{An Overview of SLMM}
As shown in the left part of Figure \ref{fig:model}, our model takes user utterance as input and conducts (K+1)-class classification to predict the corresponding class directly.
The training of our model consists of two stages: pre-training for known intents and training for open intents.
We first pre-train the model based on the labeled samples of known intents to get better intent representation.
Then we propose two strategies for training open intents, named soft labeling and manifold mixup, as shown in the middle and right part of Figure \ref{fig:model}.
Finally, our model can be directly used for inference.

\subsection{Deep Open Intent Classification Model}
We then introduce the architecture of our deep open intent classification model.
As shown in the left part of Figure \ref{fig:model}, we use BERT \cite{Devlin19} as model backbone to extract intent representation.
Given the $i$-th user utterance $x_i = \{t_1,t_2,\cdots,t_{m_i}\}$, we can get all token embeddings $[CLS,T_1,\cdots,T_{m_i}] \in \mathbb{R}^{({m_i}+1) \times H}$ from the last hidden layer of BERT, where $CLS$ is the token embedding of a special token, $m_i$ is the sentence length of the $i$-th user utterance, and $H$ is the hidden layer size.
Same as previous work \cite{Lin20,zhang21}, we apply the mean pooling layer on the token embeddings to get the averaged representation $\tilde{x}_i \in \mathbb{R}^H$:
\begin{equation}
    \tilde{x}_i = \text{mean-pooling}([CLS,T_1,\cdots,T_{m_i}])
\end{equation}

After that, we further strengthen feature extraction capability by feeding $\tilde{x}_i$ to a dense layer $h$ to get the intent representation $z_i \in \mathbb{R}^D$:
\begin{equation}
    z_i = h(x_i) = {\rm ReLU}(W_hx_i+b_h)
\end{equation}
where $D$ is the dimension of the intent representation, $W_h \in \mathbb{R}^{H \times D}$ and $b_h \in \mathbb{R}^{D}$ denote the weight and bias term of layer $h$ respectively.

\subsection{Pre-training for Known Intents}
To get better intent representation for open intent classification, we first pre-train the model based on the labeled data of known intents in training set $\mathcal{D}_{tr}$.
The intent representation $z_i$ can be learned with the softmax loss $\mathcal{L}_P$:
\begin{equation}
    \mathcal{L}_P = -\frac{1}{L} \sum_{i=1}^L {\rm log} \frac{{\rm exp}(\phi_K(z_i)^{y_i})}{\sum_{j=1}^K {\rm exp}(\phi_K(z_i)^j)}
\end{equation}
where $L$ is the size of training set, $\phi_K(\cdot)$ is a K-class classifier, and $\phi_K(\cdot)^j$ is the output logits of the $j$-th class.
Note that the classifier $\phi_K(\cdot)$ is a subset of $\phi_{K+1}(\cdot)$ and is corresponding to the classifier of the known intents part.

\subsection{Training for Open Intents}
Due to the lack of training data specific to open intents, training the model for the detection of open intents is challenging. 
To this end, we propose two strategies to generate pseudo data for the further training of model, which are soft labeling and manifold mixup.
Among these two strategies, soft labeling generate pseudo data for known intents by reshaping the label distribution of samples in training set.
Manifold mixup generates pseudo data for open intents by interpolating between the intent representations of two samples with different known intents. 

\subsubsection{Soft Labeling}

For the soft labeling strategy, our intuition is that if each training sample has a certain probability to be classified as open intents, model's overconfidence prediction on known intents would be reduced and the outlier samples of all known intents would be easier to be detected as open intents.
Thus, instead of using the one-hot label distribution, we soften the label distribution of each sample in training set by reallocating part of its probability on the class of known intents to the class of open intents.

In detail, as shown in the middle part of Figure \ref{fig:model}, for each known intent sample, we can perform soft labeling by setting the probability on open intent class as a default value $\xi$ and the probability on its ground-truth intent class as $1-\xi$.
It is worth noting that we usually set a small probability to the open class (e.g., 0.3), so that the probability of ground-truth class can be higher than that of open class.
This allows the model to have the ability of identifying open intent while at the same time avoiding the model overfitting to the open intent class.
After that, we can train the model on these pseudo samples with soft labeling via a KL-divergence loss:
 \begin{equation}
    \mathcal{L}_S = \sum_{i=1}^{L}D_{KL}\left(p(x_i)||q(x_i)\right)
\end{equation}
where $L$ is the size of training set, $p(x_i)$ is the softened probability distribution (generated by soft labeling) of the utterance $x_i$ on all intents, $q(x_i)$ is the output probability distribution of applying softmax on $\phi_{K+1}(z_i)$, and $z_i$ is the intent representation of the utterance $x_i$.

\begin{algorithm}[t]
\caption{Training Flow of SLMM}
\label{alg:inference}
\begin{algorithmic}[1]
\REQUIRE The training set
\ENSURE A open intent classification model.
\STATE \textbf{\emph{\# Pre-training for Known Intents:}}
\FORALL{iteration = 1, $\cdots$, MaxIter}
\STATE Sample a mini-batch $\{(x_i,y_i)\}$
\STATE Calculate the training loss by Eq. (3)
\STATE Obtain derivative and update the model
\ENDFOR
\STATE \textbf{\emph{\# Training for Open Intents:}}
\FORALL{iteration = 1, $\cdots$, MaxIter}
\STATE Sample a mini-batch $\{(x_i,y_i)\}$
\STATE Reshape label distribution via soft labeling
\STATE Calculate soft labeling loss by Eq. (4)
\STATE Generate pseudo data via manifold mixup
\STATE Calculate manifold mixup loss by Eq. (8)
\STATE Calculate total loss by Eq. (9)
\STATE Obtain derivative and update the model
\ENDFOR
\RETURN The model converges on validation set.
\end{algorithmic}
\end{algorithm}

\begin{table*} \setlength{\tabcolsep}{15pt}
\centering
\caption{Statistics of dataset.}\label{statis}
\begin{tabular}{lcccccc}
\toprule
\textbf{Dataset}&\textbf{Classes}&\textbf{\#Training}&\textbf{\#Validation}&\textbf{\#Test}&\textbf{Vocabulary Size}&\textbf{Length(mean)} \\
\midrule
\textbf{BAKING} & 77 &9003&1000&3080&5028&11.91  \\
\textbf{CLINC} & 150 &15000&3000&5700&8376&8.31  \\
\textbf{SNIPS} & 7 & 13084 & 700 & 700 & 11971& 9.05   \\
\textbf{ATIS} & 18 & 4978 & 500 & 893 & 938 & 11.37 \\
\bottomrule
\end{tabular}
\end{table*}

\subsubsection{Manifold Mixup}
For the manifold mixup strategy, our intuition is that if the outlier samples of known intents can be used as open intent samples for model training, the decision boundary between known intents and open intents could be better learned.
To this end, we propose to apply manifold mixup to generate open intent samples by interpolating between the representation of two samples of different known intents.
Our assumption is that the interpolated representation can be viewed as the sample of open intents.

In detail, as shown in the right part of Figure \ref{fig:model}, given two samples from different known intents, we can perform manifold mixup by interpolating their output of the $n$-th layer of BERT and labeling the interpolated representation as open intent.
Specifically, given a batch of known intent samples, we randomly select sample pairs for manifold mixup by random shuffling.
By recording the list of samples before and after random shuffling, samples at the same position in these two lists can realize random pairing.
For each sample pair $(x_i,x_j)$, if they are from different known intents (i.e., from different classes), we feed them into BERT, and get their representations until reaching $n$-th layer by:
\begin{equation}
\begin{aligned}
    h_i^l = {\rm BERT}(h_{i}^{l-1}), l\in[1,n]\\
    h_j^l = {\rm BERT}(h_{j}^{l-1}), l\in[1,n]
\end{aligned}
\end{equation}
where $h_i^l$ and $h_j^l$ are the hidden representation of sample pair $(x_i,x_j)$ after the $l$-th layer.
Then we mix the two hidden representations and continue forwarding:

\begin{equation}
    \hat{h}_m^n = \lambda h_i^n + (1-\lambda)h_j^n
\end{equation}
\begin{equation}
    \hat{h}_m^l = {\rm BERT}(\hat{h}_m^{l-1}), l\in[n+1,T]
\end{equation}
where $T$ is total number of layers in BERT, $\lambda \in$ [0,1] is a value sampled from Beta ($\alpha$,$\alpha$) distribution.
After getting the output $\hat{h}_m^T$ from BERT, we use mean-pooling and dense layer $h$ to get the intent representation $\hat{z}_m$.
The intent representation $\hat{z}_m$ is then fed into the classifier $\phi_{K+1}(\cdot)$ for open intent classification and the softmax loss is:
\begin{equation}
    \mathcal{L}_M = -\frac{1}{M} \sum_{m=1}^M {\rm log} \frac{{\rm exp}(\phi_{K+1}(\hat{z}_m)^{K+1})}{\sum_{k=1}^{K+1} {\rm exp}(\phi_{K+1}(\hat{z}_m)^k)}
\end{equation}
where and $M$ is the number of pseudo samples of open intents generated by manifold mixup, $\phi_{K+1}(\cdot)$ is the (K+1)-class linear classifier, and $\phi_{K+1}(\cdot)^k$ is the output logits of the $k$-th class.

\begin{table*}[t]\small  \setlength{\tabcolsep}{3pt}
  \centering
  \caption{Performance of open intent classification with different known class ratios (25\%, 50\%, 75\%) on four benchmark datasets. “Accuracy” and “F1” denote the accuracy score and macro F1-score over all classes. Performance (mean$\pm{\text{std}}$) over 10 runs are reported. The best results are in bold. All results of baselines on BANKING and CLINC from Zhang et al. \cite{zhang21}. The results of ADB on SNIPS and ATIS are reproduced from open-source code.}  \label{tab:performance_comparison}
    \begin{tabular}{cccccccccccc}
    \toprule
    &\multirow{2}{*}{\textbf{Methods}}&\multicolumn{2}{c}{\textbf{BANKING}}&\multicolumn{2}{c}{\textbf{CLINC}}&\multicolumn{2}{c}{\textbf{SNIPS}}&\multicolumn{2}{c}{\textbf{ATIS}}\\
    \cmidrule(lr){3-4} \cmidrule(lr){5-6} \cmidrule(lr){7-8}\cmidrule(lr){9-10} 
    &&\textbf{Accuracy}&\textbf{F1}&\textbf{Accuracy}&\textbf{F1}&\textbf{Accuracy}&\textbf{F1}&\textbf{Accuracy}&\textbf{F1}\\
    \midrule
    \multirow{6}{1cm}{\tabincell{c}{\textbf{25\%}}}&
    \textbf{MSP}&43.67&50.09&47.02&47.62&-&-&-&-\\
    &\textbf{DOC}&56.99&58.03&74.97&66.37&-&-&-&-\\
    &\textbf{OpenMax}&49.94&54.14&68.50&61.99&-&-&-&-\\
    &\textbf{LMLC}&64.21&61.36&81.43&71.16&-&-&-&-\\
    &\textbf{ADB}&78.85&71.62&87.59&77.19&67.41&69.89&63.90&53.21&\\
    &\textbf{SLMM}&\bf{80.96$\pm{\bf{0.83}}$}&\bf{72.58$\pm{\bf{0.79}}$}&\bf{91.25$\pm{\bf{0.25}}$}&\bf{80.04$\pm{\bf{1.22}}$}&\bf{75.30$\pm{\bf{4.72}}$}&\bf{74.64$\pm{\bf{3.48}}$}&\bf{69.90$\pm{\bf{10.81}}$}&\bf{61.18$\pm{\bf{9.44}}$}\\
    \midrule
    \midrule
    \multirow{6}{1cm}{\tabincell{c}{\textbf{50\%}}}&
    \textbf{MSP}&59.73&71.18&62.96&70.41&-&-&-&-\\
    &\textbf{DOC}&64.81&73.12&77.16&78.26&-&-&-&-\\
    &\textbf{OpenMax}&65.31&74.24&80.11&80.56&-&-&-&-\\
    &\textbf{LMLC}&72.73&77.53&83.35&82.16&-&-&-&-\\
    &\textbf{ADB}&78.86&80.90&86.54&85.05&80.36&82.96&81.27&63.24\\
    &\textbf{SLMM}&\bf{80.28$\pm{\bf{0.57}}$}&\bf{81.98$\pm{\bf{0.47}}$}&\bf{89.04$\pm{\bf{0.27}}$}&\bf{86.72$\pm{\bf{0.08}}$}&\bf{80.93$\pm{\bf{0.08}}$}&\bf{83.15$\pm{\bf{0.15}}$}&\bf{87.37$\pm{\bf{2.69}}$}&\bf{70.21$\pm{\bf{2.44}}$}\\
    \midrule
    \midrule
    \multirow{6}{1cm}{\tabincell{c}{\textbf{75\%}}}&
    \textbf{MSP}&75.89&83.60&74.07&82.38&-&-&-&-\\
    &\textbf{DOC}&76.77&83.34&78.73&83.59&-&-&-&-\\
    &\textbf{OpenMax}&77.45&84.07&76.80&73.16&-&-&-&-\\
    &\textbf{LMLC}&78.52&84.31&83.71&86.23&-&-&-&-\\
    &\textbf{ADB}&81.08&85.96&86.32&88.53&82.56&85.13&87.53&71.60\\
    &\textbf{SLMM}&\bf{82.18$\pm{\bf{1.42}}$}&\bf{86.67$\pm{\bf{0.75}}$}&\bf{88.88$\pm{\bf{0.49}}$}&\bf{90.40$\pm{\bf{0.53}}$}&\bf{85.37$\pm{\bf{0.14}}$}&\bf{85.57$\pm{\bf{0.21}}$}&\bf{94.33$\pm{\bf{0.56}}$}&\bf{81.00$\pm{\bf{1.03}}$}\\
    \bottomrule
    \end{tabular}
\end{table*}

\subsubsection{Overall Training Objective}
Based on two kinds of pseudo data generated by soft labeling and manifold mixup, we can get the final training objective by combining the soft labeling loss and manifold mixup loss:
\begin{equation}
    \mathcal{L} = \mu\mathcal{L}_S + (1-\mu) \mathcal{L}_M
\end{equation}
where $\mu$ is a tradeoff parameter.

\subsection{Training Flow and Inference}
In summary, there are two steps in our method to train the deep open intent classification model, i.e., first pre-training for known intents on the original training set, and then training for open intents.
The whole training flow of our method is illustrated in Algorithm \ref{alg:inference}.

In the inference phase, we get the utterance feature and use the classifier $\phi_{K+1}(\cdot)$ to get the predicted class.

\section{Experiments}

\subsection{Datasets and Evaluation Metrics}
To verify the effectiveness of our model, we conduct experiments on four benchmark datasets, which include BANKING, CLINC, SNIPS, and ATIS.

\textbf{BANKING} is a dataset containing 77 intents and 13,083 customer service queries in the banking domain \cite{banking}.

\textbf{CLINC} is a dataset containing 22,500 in-scope queries covering 150 intents and 1,200 out-of-scope queries across 10 domains\cite{DBLP:conf/emnlp/LarsonMPCLHKLLT19}.


\textbf{SNIPS} is a dataset containing 7 intents across different domain \cite{DBLP:journals/corr/abs-1805-10190}.

\textbf{ATIS} is a dataset containing 18 intents in the airline travel domain \cite{DBLP:conf/naacl/HemphillGD90}.

For BANKING and CLINC dataset, we use the processed data provided by Zhang et al. \cite{zhang21}.
And for SNIPS and ATIS dataset, we use the processed data provided by Lin et al. \cite{Lin19}.
The detailed statistics of four datasets are shown in Table \ref{statis}.

Following previous studies \cite{Shu17,Yan20,Lin19,zhang21}, we also use accuracy score (Accuracy) and macro F1-score (F1) as evaluation metrics for performance measuring.
Besides the overall Accuracy and F1, we also report macro F1-score over known intent classes and open intent class.
We use the open intent class to represent all unknown intents.
\begin{table*}[t]\small  \setlength{\tabcolsep}{3pt}
  \centering
  \caption{Performance of open intent classification with different known class ratios (25\%, 50\%, 75\%) on four benchmark datasets. “Open” and “Known” denote the macro F1-score over open class and known classes respectively. APerformance (mean$\pm{\text{std}}$) of our method over 10 runs are reported. The best results are in bold. All results of baselines on BANKING and CLINC from Zhang [10]. The results of ADB on SNIPS and ATIS are reproduced from open-source code.}  \label{tab:open}
    \begin{tabular}{ccccccccccccc}
    \toprule
    &\multirow{2}{*}{\textbf{Methods}}&\multicolumn{2}{c}{\textbf{BANKING}}&\multicolumn{2}{c}{\textbf{CLINC}}&\multicolumn{2}{c}{\textbf{SNIPS}}&\multicolumn{2}{c}{\textbf{ATIS}}\\
    \cmidrule(lr){3-4} \cmidrule(lr){5-6}  \cmidrule(lr){7-8} \cmidrule(lr){9-10}
    &&\textbf{Open}&\textbf{Known}&\textbf{Open}&\textbf{Known}&\textbf{Open}&\textbf{Known}&\textbf{Open}&\textbf{Known}\\
    \midrule
    \multirow{6}{1cm}{\tabincell{c}{\textbf{25\%}}}&
    \textbf{MSP}&41.43&50.55&50.88&47.53&-&-&-&-\\
    &\textbf{DOC}&61.42&57.85&81.98&65.96&-&-&-&-\\
    &\textbf{OpenMax}&51.32&54.28&75.76&61.62&-&-&-&-\\
    &\textbf{LMLC}&70.44&60.88&87.33&70.73&-&-&-&-\\
    &\textbf{ADB}&84.56&70.94&91.84&76.80&70.99&69.34&64.93&50.33\\
    &\textbf{SLMM}&\bf{86.53$\pm{\bf{0.51}}$}&\bf{71.85$\pm{\bf{0.84}}$}&\bf{94.53$\pm{\bf{0.62}}$}&\bf{79.66$\pm{\bf{2.81}}$}&\bf{79.68$\pm{\bf{4.67}}$}&\bf{72.13$\pm{\bf{1.51}}$}&\bf{67.14$\pm{\bf{11.79}}$}&\bf{59.84$\pm{\bf{7.42}}$}\\
    \midrule
    \midrule
    \multirow{6}{1cm}{\tabincell{c}{\textbf{50\%}}}&
    \textbf{MSP}&41.19&71.97&57.62&70.58&-&-&-&-\\
    &\textbf{DOC}&55.14&73.59&79.00&78.25&-&-&-&-\\
    &\textbf{OpenMax}&54.33&74.76&81.89&80.54&-&-&-&-\\
    &\textbf{LMLC}&69.53&77.74&85.85&82.11&-&-&-&-\\
    &\textbf{ADB}&78.44&80.96&88.65&85.00&74.92&84.97&77.27&61.40\\
    &\textbf{SLMM}&\bf{80.14$\pm{\bf{0.40}}$}&\bf{82.03$\pm{\bf{0.56}}$}&\bf{91.07$\pm{\bf{0.05}}$}&\bf{86.67$\pm{\bf{1.11}}$}&\bf{75.07$\pm{\bf{2.76}}$}&\bf{85.17$\pm{\bf{0.08}}$}&\bf{85.66$\pm{\bf{3.41}}$}&\bf{68.05$\pm{\bf{1.79}}$}\\
    \midrule
    \midrule
    \multirow{6}{1cm}{\tabincell{c}{\textbf{75\%}}}&
    \textbf{MSP}&39.23&84.36&59.08&82.59&-&-&-&-\\
    &\textbf{DOC}&50.60&83.91&72.87&83.69&-&-&-&-\\
    &\textbf{OpenMax}&50.85&84.64&76.35&73.13&-&-&-&-\\
    &\textbf{LMLC}&58.54&84.75&81.15&86.27&-&-&-&-\\
    &\textbf{ADB}&66.47&86.29&83.92&88.58&69.30&88.29&71.08&71.72\\
    &\textbf{SLMM}&\bf{69.20$\pm{\bf{0.50}}$}&\bf{86.97$\pm{\bf{1.08}}$}&\bf{87.15$\pm{\bf{1.07}}$}&\bf{90.43$\pm{\bf{0.32}}$}&\bf{69.72$\pm{\bf{2.68}}$}&\bf{89.00$\pm{\bf{0.15}}$}&\bf{79.84$\pm{\bf{2.68}}$}&\bf{81.08$\pm{\bf{0.92}}$}\\
    \bottomrule
    \end{tabular}
\end{table*}

\begin{table}\small  \setlength{\tabcolsep}{1.5pt}
  \centering
  \caption{Ablation study of open intent classification with different known class ratios (25\%, 50\%, 75\%) on CLINC dataset. F1 denotes macro-F1 over all classes. Acc-KoK denotes the accuracy of known intents classifier on samples of known intents. Averaged results over 10 runs are reported. The best results are in bold.}  \label{tab:aba}
    \begin{tabular}{cccccccc}
    \toprule
    &\multirow{2}{*}{\textbf{Setting}}&\multicolumn{4}{c}{\textbf{CLINC}}\\
    \cmidrule(lr){3-6}
    &&\textbf{Accuracy}&\textbf{F1}&\textbf{Weighed-F1}&\textbf{Acc-KoK}\\
    \midrule
    \multirow{5}{1cm}{\tabincell{c}{\textbf{25\%}}}&
    SLMM&\bf{91.25}&80.04&\bf{90.97}&\bf{97.81}\\
    &w/o pre-training&84.01&69.30&88.94&95.18\\
    &w/o SL&84.78&74.82&81.55&95.35\\
    &w/o MM&89.84&\bf{80.26}&90.36&97.54\\
    &w/o SL\&MM&19.78&37.31&8.18&97.11\\
    \midrule
    \midrule
    \multirow{5}{1cm}{\tabincell{c}{\textbf{50\%}}}&
    SLMM&\bf{89.04}&86.72&\bf{89.45}&96.93\\
    &w/o pre-training&81.22&76.48&87.86&95.47\\
    &w/o SL&63.84&71.44&72.76&90.98\\
    &w/o MM&88.35&\bf{86.74}&89.00&\bf{97.11}\\
    &w/o SL\&MM&39.95&59.92&26.09&96.84\\
    \midrule
    \midrule
    \multirow{5}{1cm}{\tabincell{c}{\textbf{75\%}}}&
    SLMM&\bf{88.88}&\bf{90.40}&\bf{88.70}&\bf{96.40}\\
    &w/o pre-training&81.21&81.35&88.07&95.83\\
    &w/o SL&72.14&81.22&81.01&91.16\\
    &w/o MM&88.25&90.14&88.46&96.25\\
    &w/o SL\&MM&59.25&75.05&46.98&96.01\\
    \bottomrule
    \end{tabular}
\end{table}

\subsection{Baselines}
We compare our method with the following state-of-the-art open intent classification methods:
\begin{itemize}
\item \textbf{OpenMax} uses OpenMax to replace the softmax layer and then uses the Weibull distribution to calibrate it \cite{Bendale16}.
\item \textbf{MSP} uses the maximum prediction probability as confidence score to detect whether this example belongs to unknown intents \cite{Hendrycks17}. We use the same confidence threshold (i.e., 0.5) as in \cite{Lin19,zhang21}.
\item \textbf{DOC} replaces softmax layer with sigmoid activation function as the final layer and uses a statistics approach to determine the threshold \cite{Shu17}.
\item \textbf{LMCL} uses margin loss to learn discriminative features and then uses local outlier factor algorithm to detect unknown intents \cite{Lin19}.
\item \textbf{ADB} uses a post-processing method to learn the adaptive spherical decision boundaries \cite{zhang21}.
\end{itemize}
To make a fair comparison, BERT used in our model is adopted as model backbone of all baselines.

\subsection{Experimental Settings}
Following the same settings as in previous studies \cite{Shu17,Lin19,zhang21}, all datasets are divided into training, validation and test sets.
The number of known classes are varied with the proportions of 25\%, 50\%, and 75\% in the training set.
The remaining classes are regarded as open class and removed from the training set.
In the testing phase, both known classes and open class are used. 
For each experimental setting, we report the average performance over ten runs of experiments.

We adopt BERT-Base \cite{Devlin19} as model backbone in our work.
To speed up the training procedure and achieve better performance, we freeze all but the last transformer layer of BERT.
For soft labeling, the default probability on open intent class is $\xi = 0.3$.
For manifold mixup, we interpolate the hidden state before the last transformer layer of BERT and set $\alpha = 2$ for the beta distribution.
Besides, we use AdamW \cite{DBLP:conf/iclr/LoshchilovH19} as the optimizer and initialize the learning rate and batch size to 2e-5 and 128 respectively.
During training, we linear warmup the learning rate at the first 10\% of all training steps and then linear decay is used.
We train our model for 100 epochs and select the best model based on the performance on the validation set with early stopping.

\subsection{Results}

Table \ref{tab:performance_comparison} and Table \ref{tab:open} show the performance of our proposed method and baseline methods for the open intent classification task on four datasets.
Table \ref{tab:performance_comparison} shows the overall performance (i.e., accuracy score and macro F1-score over all classes).
Table \ref{tab:open} shows the fine-grained performance (i.e., macro F1-score over open class and known classes).

First, by observing overall performance in Table \ref{tab:performance_comparison}, our method consistently outperforms all baselines with 25\%, 50\% and 75\% known intent classes on four datasets and most standard deviation values of our method are smaller than the performance gap between our method and the previous state-of-the-art method ADB.
Compared with the best baseline method ADB on accuracy, our method outperforms it by 3.66\%, 2.5\%, and 2.56\% on CLINC dataset, by 2.11\%, 1.42\%, and 1.1\% on BANKING dataset with 25\%, 50\%, and 75\% settings respectively.
This shows the effectiveness of our method.
When the ratio of known intent classes is low, the performance of most baselines is poor, and our method still performs well and achieves robust results with fewer training samples.
And the advantages of our method are more obvious when the ratio of known intent classes is low.
It is worth noting that ATIS dataset is imbalanced dataset, so when the ratio of known intent classes is 25\%, ADB and SLMM perform poorly. 

Second, by observing the fine-grained performance in Table \ref{tab:open}, we can find that our method consistently outperforms all baselines on open class and known classes.
Especially on CLINC dataset, our model outperforms the best baseline method ADB by 2.69\%, 2.42\%, and 3.23\% on open class and 2.86\%, 1.67\%, and 1.85\% on known classes with 25\%, 50\%, and 75\% settings, respectively.
The advantages of our method are more obvious for open class.
This shows that our proposed method is not only effective for detecting open class but also can better classify known classes.

\begin{figure*}
  \centering
  \subfigure[Effects of the default probability $\xi$ on open intent class.]{
        \includegraphics [width=0.235 \textwidth]{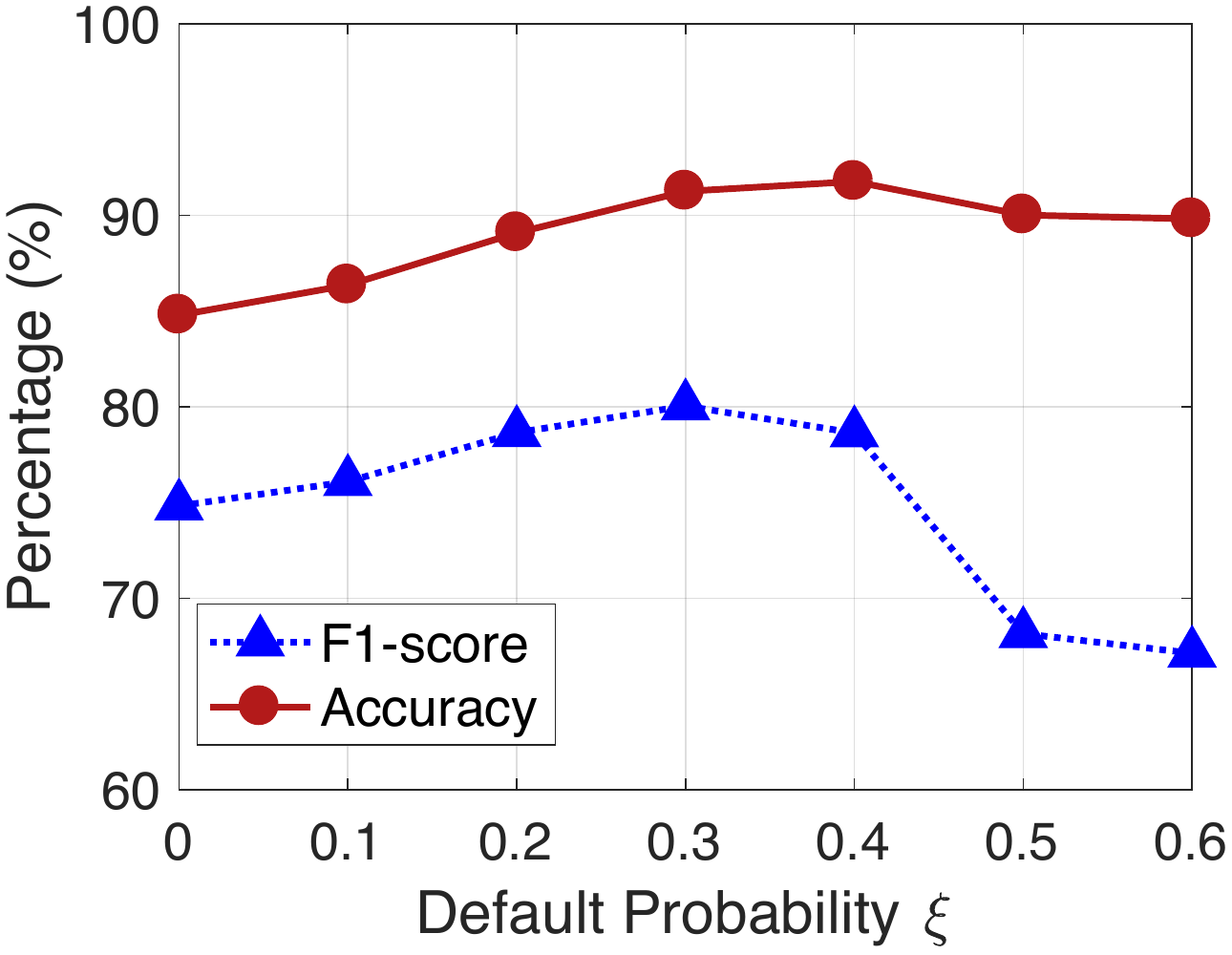}
 \includegraphics [width=0.235 \textwidth]{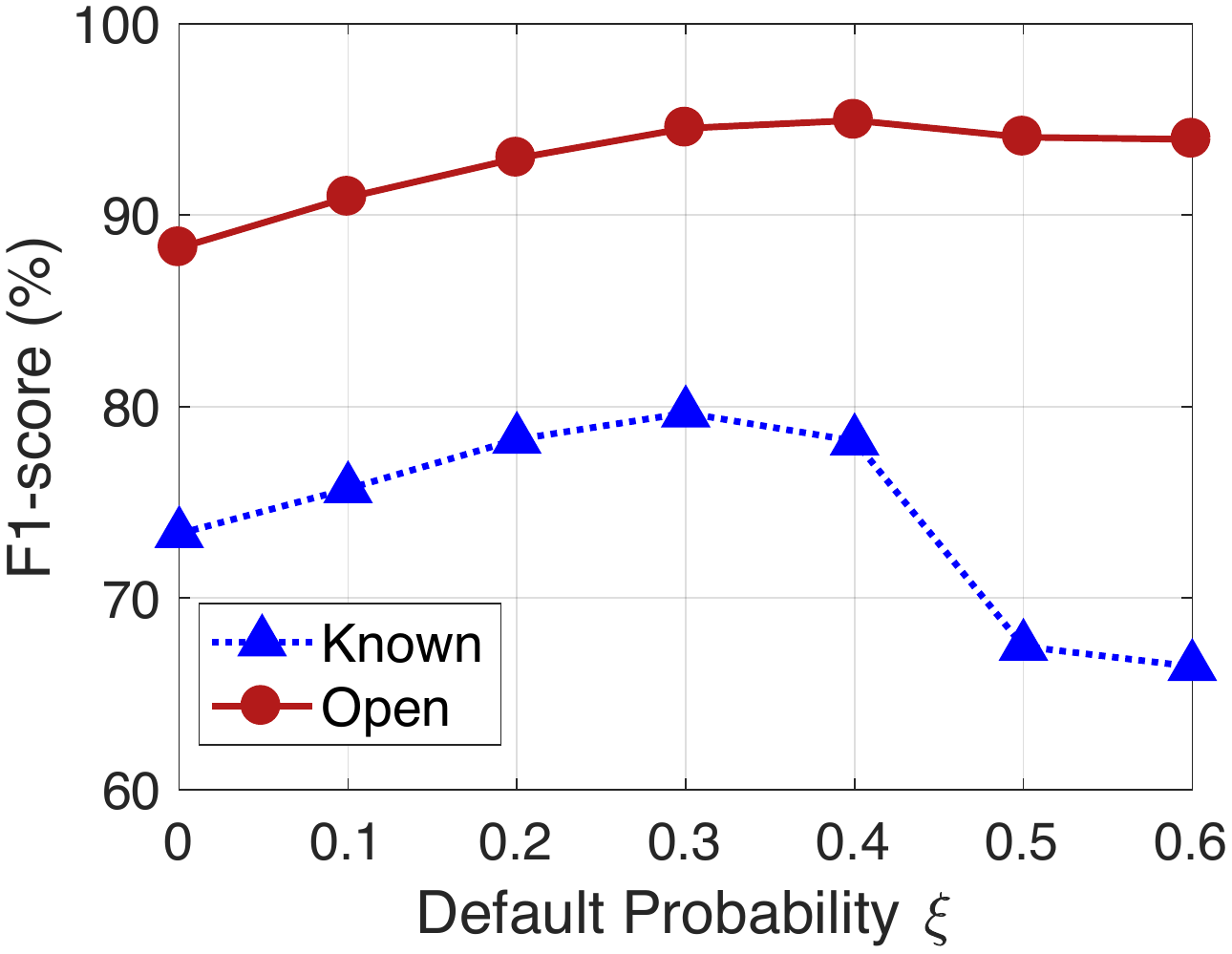}
  }
  \subfigure[Effects of the tradeoff parameter $\mu$ in loss function.]{
         \includegraphics [width=0.235 \textwidth]{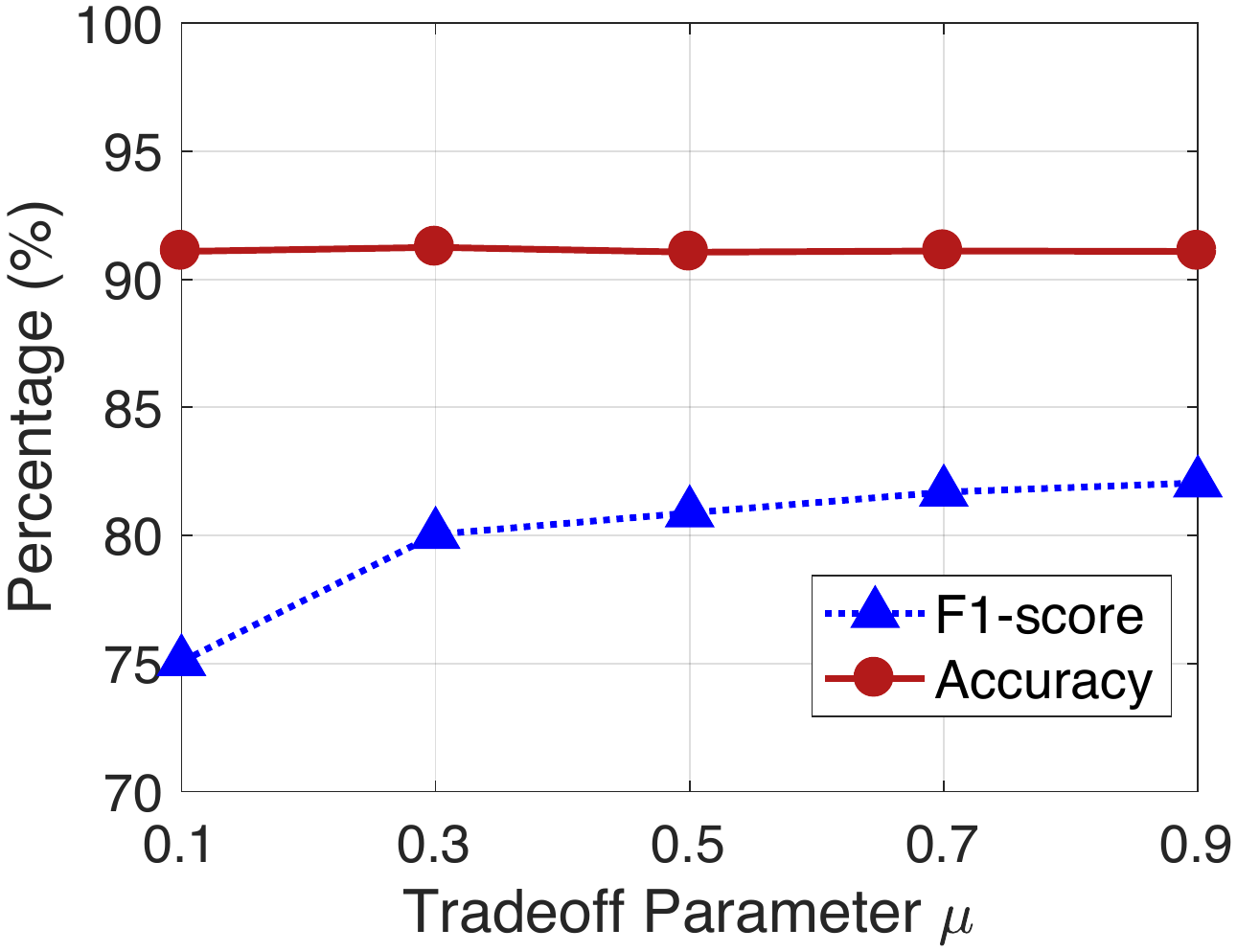}
  \includegraphics [width=0.235 \textwidth]{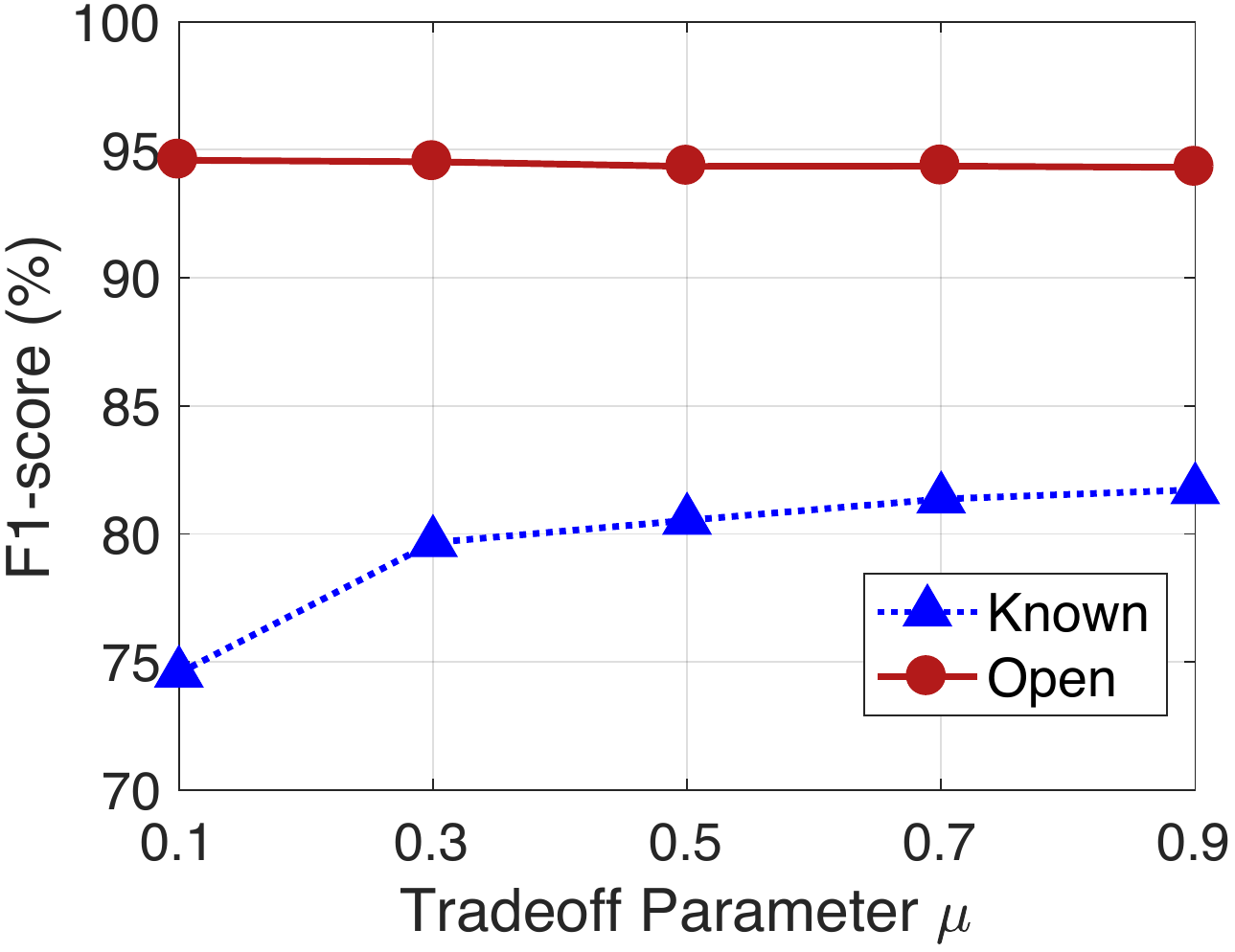}
  }
  \caption{Effects of $\xi$ and $\mu$ on CLINC dataset with 25\% known classes. Left part is the effect of $\xi$ and the right part is the effect of $\mu$.} \label{hyper}
\end{figure*}

\subsection{Ablation Study}
In this section, we conduct experiments to explore the effects of pre-training, soft labeling, and manifold mixup on our model.
To conduct a comprehensive analysis, we report the performance on four metrics: accuracy, F1, weighted-F1, and Acc-KoK.
Among these metrics, weighted-F1 is mainly used to show the impact of class imbalance, and Acc-KoK is mainly used to analyze the impact of each component on the known intent classification under the closed-world setting.

First, as shown in Table \ref{tab:aba}, by observing the overall performance on accuracy, macro-F1, and weighted-F1, we can find that the performance drops after removing each of the three components (i.e., pre-training, SL, and MM) individually.
This indicates that all components contribute to the final performance.

Second, by observing the performance of removing pre-training, we can find that the accuracy drops by about 7\%, 8\%, and 7\% with three settings respectively.
This indicates that pre-training for known intents can learn better intent representation and is necessary for the training of SLMM.

Third, by observing the performance of removing soft labeling (implemented by setting $\xi = 0$), we can find that the accuracy of removing soft labeling drops by 6.47\%, 25.2\%, and 16.1\% with three settings respectively.
This shows that soft labeling is effective.

Fourth, by observing the performance of removing manifold mixup (i.e., w/o MM), we can find that the accuracy and weighted-F1 consistently decreases, but the macro-F1 rises a little when the known intent ratio is 25\% and 50\%.
Such inconsistency trend of metrics is mainly caused by the class imbalance.
Since MM strategy only generate pseudo samples of open intents, MM makes the model more inclined to improve the performance on open intent class while degrading the performance on known intent classes.
Thus, when the ratio of open intents is large but the class weights are the same, the improvement on open intent class can not be effectively reflected by macro-F1.
From the perspective of accuracy and weighted-F1, we can find that MM can further improve the performance of our method upon the using of SL.
When the amount of data is more (i.e., 75\% known classes), macro F1-score over all classes drops.

Fifth, we can see that the model performs poorly after removing both soft labeling and manifold mixup.
This is because after removing soft labeling and manifold mixup, the model is only trained for known classes and cannot identify the open class.

\begin{figure*}
  \centering
        \includegraphics [width=0.3 \textwidth]{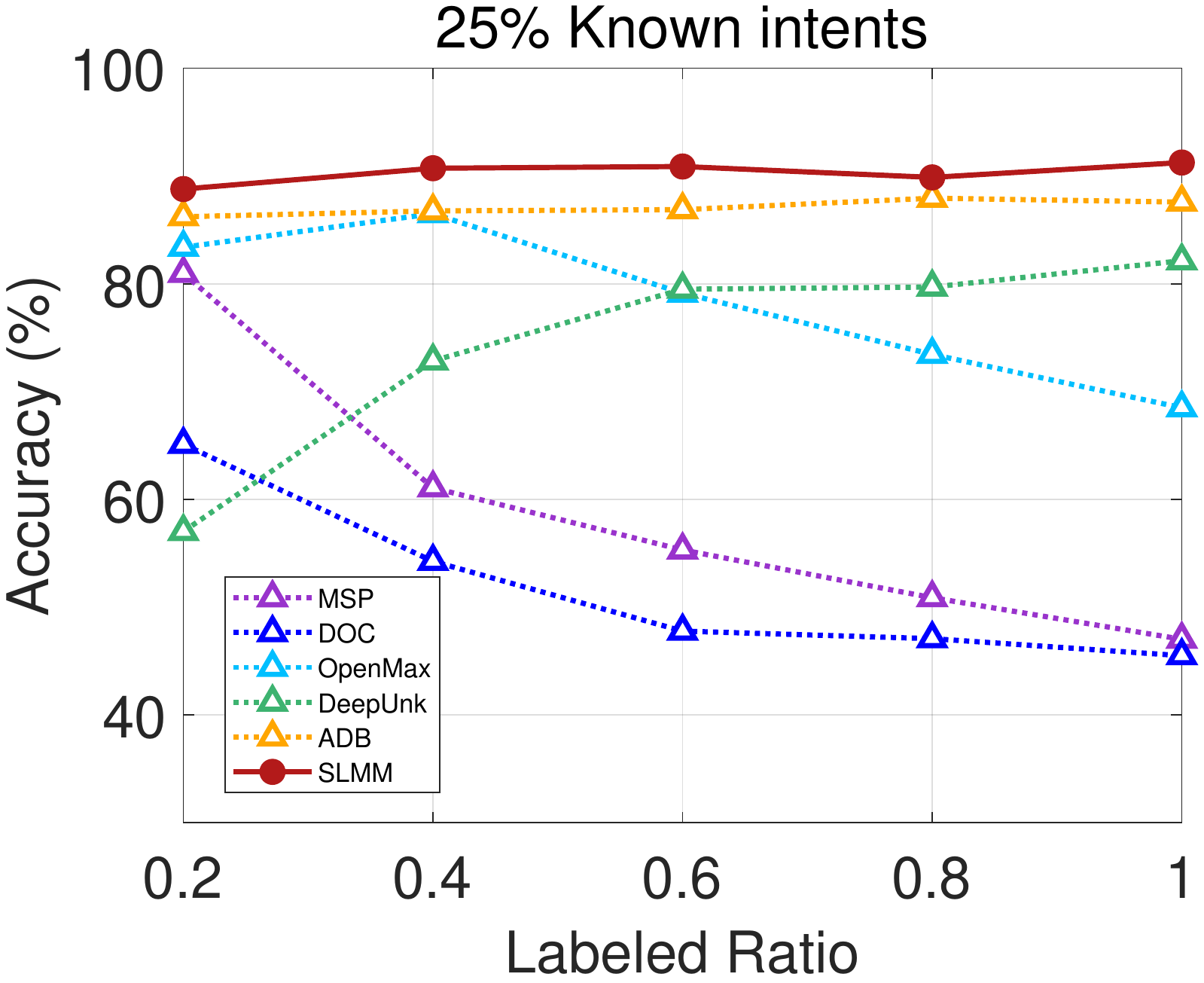}
        \hspace{2mm}
 \includegraphics [width=0.3 \textwidth]{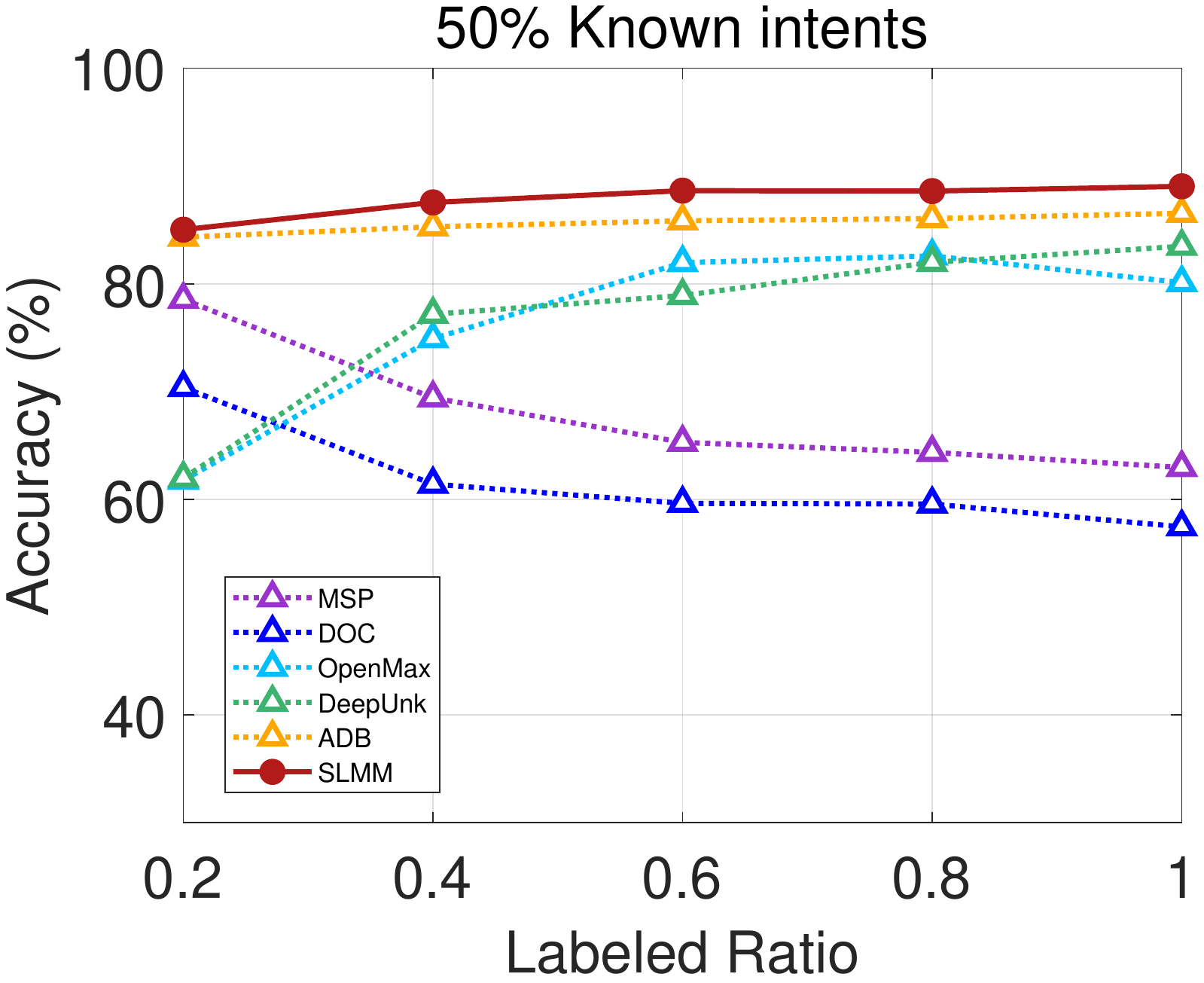}
 \hspace{2mm}
 \includegraphics [width=0.3 \textwidth]{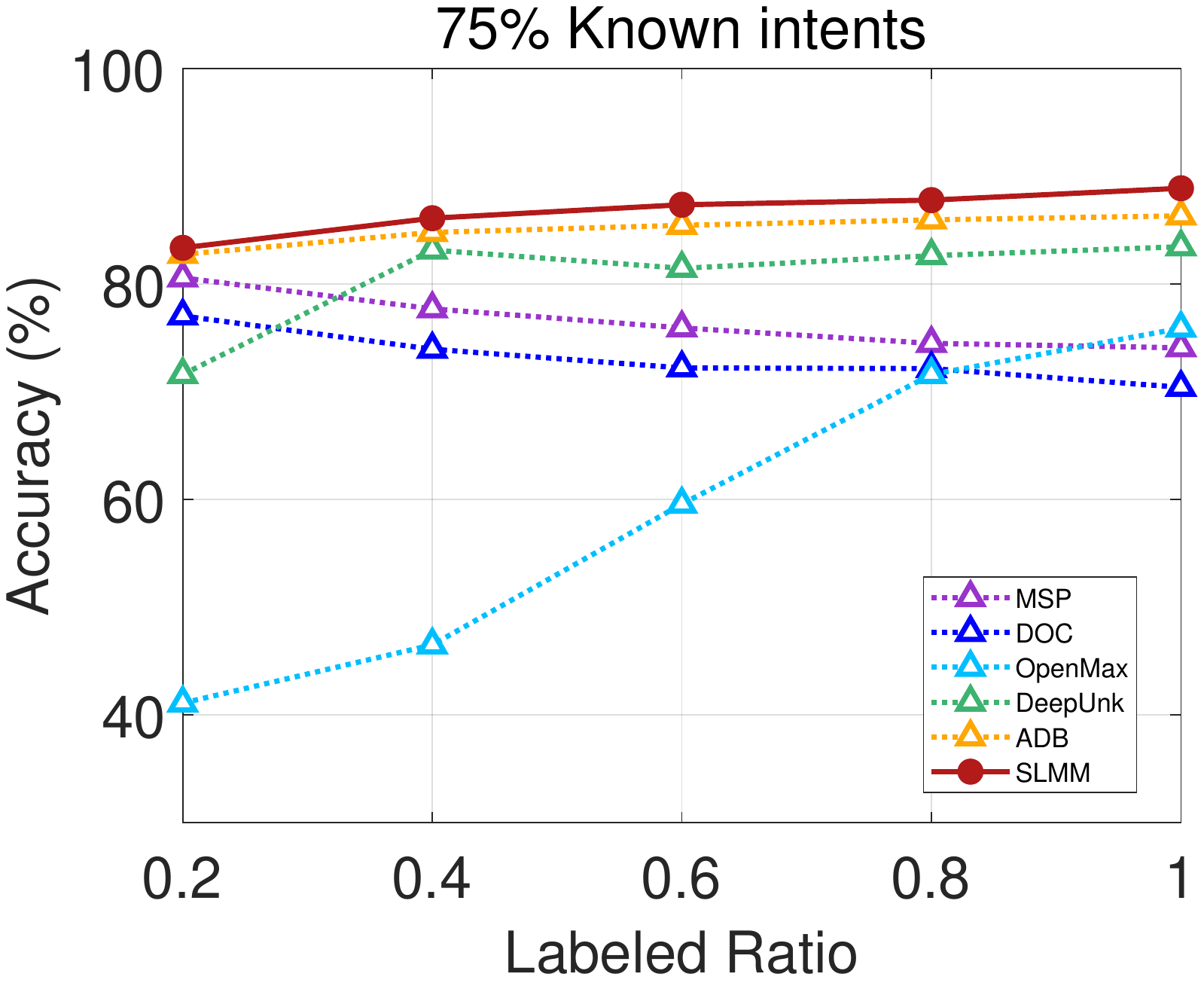}
  \caption{Effects of the labeled ratio on CLINC dataset with different known class proportions (25\%, 50\%, 75\%).} \label{labeled}
\end{figure*}

Finally, we also list the accuracy of the known intent classifier on samples of known intents (i.e., Acc-KoK). 
We can find that using both SL and MM (i.e., SLMM) does not degrade the performance of the pre-trained konwn intent classification model (i.e., w/o SL\&MM), but using MM alone (i.e., w/o SL) degrades the performance of pre-trained model. 
This implies that SL is very important to maintain the performance of the pre-trained model on samples of known intents.
The reason may be that the soft labeling strategy plays a role similar to label smoothing, thus it can help to improve the generalization performance of the model.

\subsection{Model Analysis}

\subsubsection{Effect of Soft Labeling}
In this part, we explore the effect of $\xi$ on the performance of our method, which controls the default probability of open class in soft labeling.
We conduct experiment by varying $\xi$ from 0 to 0.6 step by 0.1 on CLINC dataset with 25\% known classes.
As shown in Figure \ref{hyper}(a), the overall performance increases first and then decreases with varied $\xi$.
This is because as $\xi$ increases, the model can effectively alleviate overconfident prediction problem and identify open class.
When $\xi$ is equal to 0.3, the performance (i.e., F1-score) is best.
Then the bigger $\xi$ has a negative effect on predicting known classes and the performance decreases.
This is because a large $\xi$ will make the model more inclined to predict samples as open class, thus the F1-score of known classes will drop significantly.

\begin{table} \setlength{\tabcolsep}{10pt}
\centering
\caption{Effect of $\alpha$ of Beta distribution in manifold mixup.}\label{alpha}
\begin{tabular}{ccccccc}
\toprule
{\bm \alpha}&\textbf{0.5}&\textbf{1}&\textbf{2}&\textbf{4}\\
\midrule
\textbf{Accuracy} &81.50&90.22&91.25&\bf{91.85} \\
\textbf{F1}&15.08&69.78&\bf{80.04}& 79.79\\
\bottomrule
\end{tabular}
\end{table}

\begin{table} \setlength{\tabcolsep}{15pt}
\centering
\caption{Effect of interpolation position $n$ in manifold mixup.} \label{n}
\begin{tabular}{ccccccc}
\toprule
\textbf{n}&\textbf{9}&\textbf{10}&\textbf{11}\\
\midrule
\textbf{Accuracy} &87.03&86.85&\bf{91.25} \\
\textbf{F1}&75.30&75.18&\bf{80.04}\\
\bottomrule
\end{tabular}
\end{table}

\subsubsection{Effect of Manifold Mixup}
In this part, we explore the effect of $\alpha$ and $n$ on the performance of our method, which are hyperparameters of Beta distribution and interpolation position in manifold mixup.
We conduct experiment by varying $\alpha$ (i.e., 0.5, 1, 2, and 4) and $n$ (i.e., 9, 10, and 11) on CLINC dataset with 25\% known classes.
First, we explore the effect of $\alpha$.
When $\alpha$ equals to 0.5, the sampled $\lambda$ from Beta distribution is near 0 or 1.
When $\alpha$ equals to 1, the sampled $\lambda$ is uniform.
When $\alpha$ equals to 2 or 4, the sampled $\lambda$ is close to 0.5.
As shown in Table \ref{alpha}, $\alpha$ has a significant impact on performance.
When $\alpha$ is equal to 0.5 or 1, the model does not perform well.
This is because the quality of the generated samples is poor, and their representation is close to samples of known classes (i.e., high-confidence areas).
This makes the model confused about distinguishing between samples of known classes and open class. 
When $\alpha$ is equal to 2 or 4, the model performs well.
This is due to the generated samples are in the low-confidence areas, and these samples can help model to learn a better decision boundary.
When $\alpha$ is equal to 2 or 4, the overall performance does not change much.
So we set $\alpha$ = 2 by default.
Second, we explore the effect of $n$.
From Table~\ref{n}, we can find that when $n$ equals to 11, model achieves the best performance.
The performance of setting smaller $n$ is worse than that of setting $n=11$.
This may be because we fix all the parameters of BERT but the last layer and previous fixed transformer layers cannot effectively adapt to the change of representation mode brought by manifold mixup.

\begin{figure}
\centering
\includegraphics[width=0.8\columnwidth]{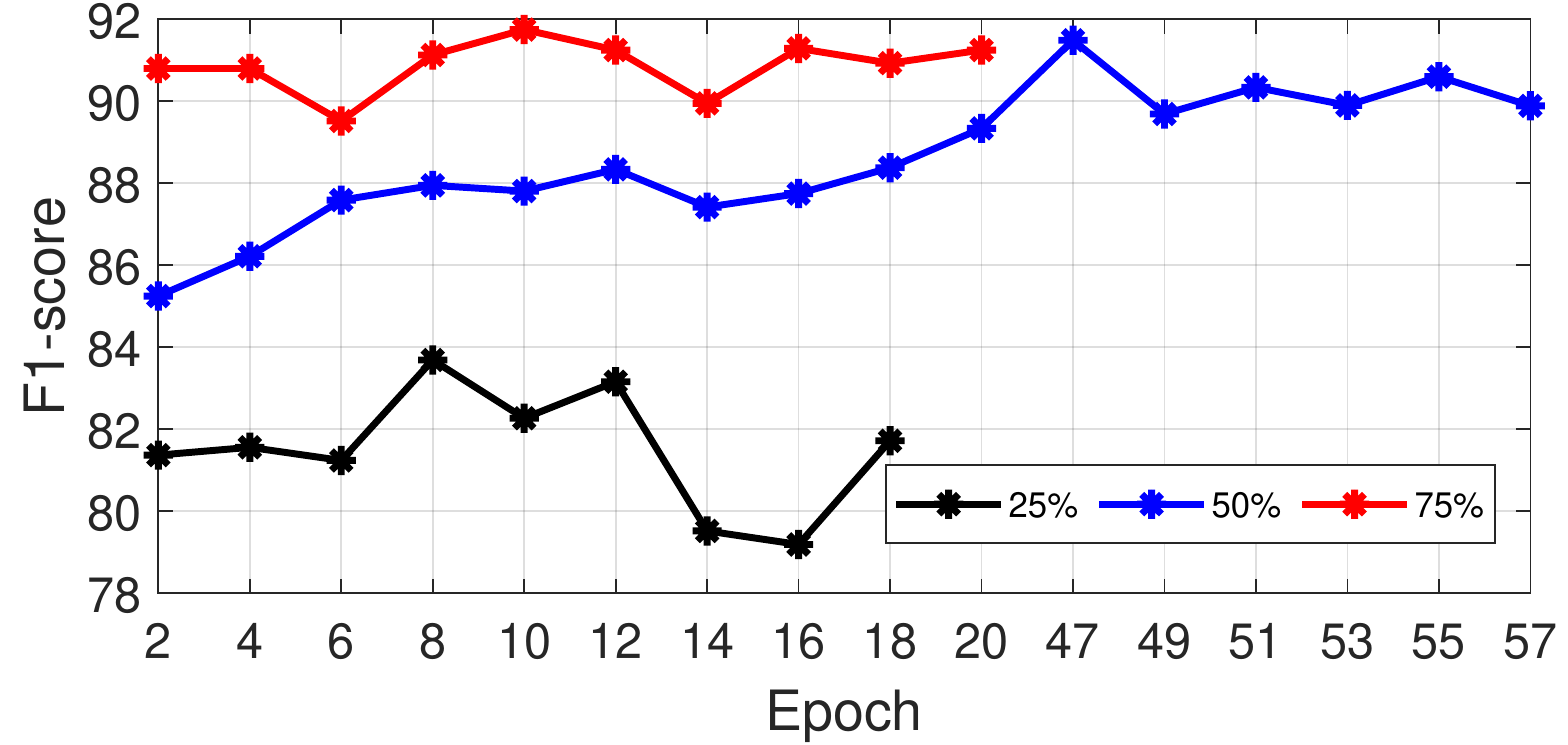}
\caption{The performance of our method on the validation set of CLINC dataset with the increase of training epoch.} \label{learning_process}
\end{figure}

\subsubsection{Effect of Tradeoff Parameter}
In this part, we explore the effect of the tradeoff parameter $\mu$ on the performance of our method.
We conduct experiment by varying $\mu$ from 0.1 to 0.9 step by 0.2 on CLINC dataset with 25\% known classes.
The results are shown in Figure \ref{hyper}(b).
We can observe that SLMM achieves the relatively stable performance on accuracy with varied $\mu$, which indicates the robustness of our method.
When $\mu = 0.3$, we can get the best accuracy.
As $\mu$ increases, macro F1-score over open class decreases slightly and macro F1-score over all classes and known classes rises.
This is because as $\mu$ increases, model will pay less attention to pseudo samples generated through manifold mixup and is more inclined to predict samples as known intent classes.
And macro F1-score over all classes is closer to macro F1-score over known classes.
Therefore, macro F1-score over all classes and known classes have an upward trend, and macro F1-score over open class has a downward trend.

\subsubsection{Effect of Labeled Data}
In this part, we explore the effect of labeled data by varying the labeled ratio in the range of 0.2, 0.4, 0.6, 0.8, and 1.0 on CLINC dataset with 25\%, 50\%, and 75\% known intent classes.
First, we can see that our model achieves the best performance on all settings as shown in Figure \ref{labeled}.
This shows the effectiveness of our method.
And the advantage of our method is more obvious, when the amount of labeled data is small with 25\% known intents.
Second, we can see that the performance of our model is stable, and the statistic-based methods (i.e., MSP and DOC) only work well with less labeled data and perform poorly as the ratio of labeled data increases.
This is because these methods tend to be biased towards the known intent classes as the number of labeled data increases.
Some deep learning methods (i.e., OpenMax and DeepUnk) perform poorly with less labeled data.
This is because the centroids and representations learned by their method are biased with less labeled data.
Both our method and ADB achieve stable performance, but our method achieves a better performance than ADB.

\subsubsection{Effect of Training Epoch}
We further explore how the performance of our method varies on validation set with the increase of training epoch.
As shown in Figure~\ref{learning_process}, three lines are drawn corresponding to three settings of known intent ratio (i.e., 25\%, 50\%, and 75\%) on the CLINC dataset.
To avoid endless training, we set the maximum training epoch as 100, but early stop the training if the best performance on validation set is not updated for 10 consecutive epochs.
As shown, the lines of 25\%, 50\%, and 75\% complete the training process with early stopping at the 18th, 57th, and 20th epoch respectively.
The best performance of our method on validation set is thus achieved at the 8th, 47th, and 10th epoch respectively for the three settings.
The fact that so many epochs are needed to retrain the model implies that the SLMM phase made a large adjustment to the representation space of the pre-training step.

\begin{figure}[t]
\centering
\includegraphics[width=0.5\columnwidth]{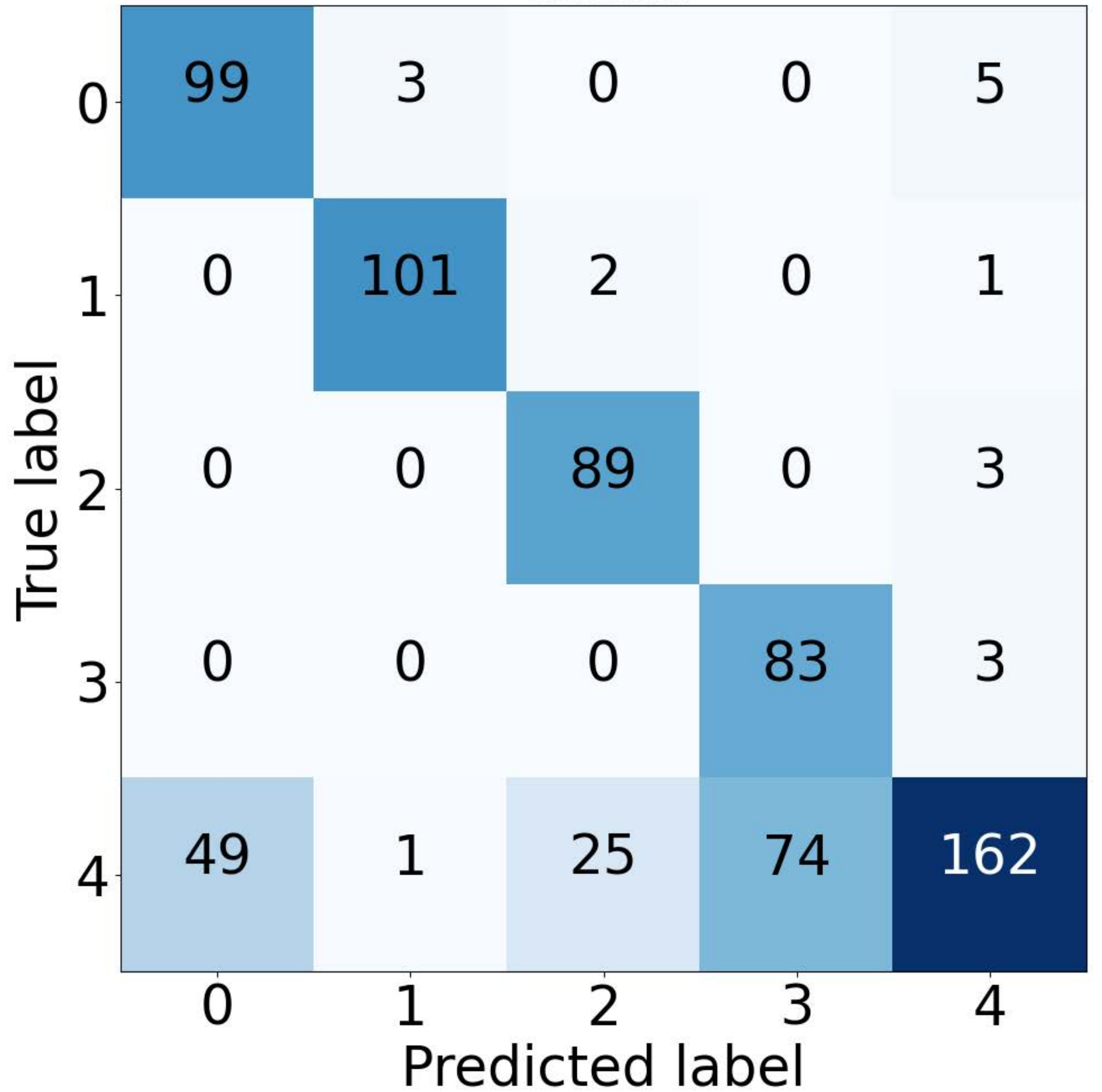}
\caption{Confusion matrix of our model on SNIPS dataset with 50\% known classes. Class 4 denotes the open class.}\label{cm}
\end{figure}

\begin{table}\scriptsize\setlength{\tabcolsep}{1pt}
\centering
\caption{Two examples for error analysis.} \label{case_study}
\begin{tabular}{l|c|c}
\toprule
\textbf{Utterance} & \textbf{Ground-Truth Intent} &\textbf{Predicted Intent}\\
\midrule
Add this song to blues roots. & \tabincell{c}{Add To Play List \\ \emph{\textbf{Open intents}}} & \tabincell{c}{Play Music \\ \emph{\textbf{Known intents}}}\\
\midrule
Find a movie called living in america. & \tabincell{c}{Search Creative Work \\ \emph{\textbf{Open intents}}} &  \tabincell{c}{Search Screening Event \\ \emph{\textbf{Known intents}}}\\
\bottomrule
\end{tabular}
\end{table}

\subsection{Error Analysis}

\begin{figure*}
  \centering
    \subfigure[50\% known classes on SNIPS.]{
         \includegraphics [width=0.27 \textwidth]{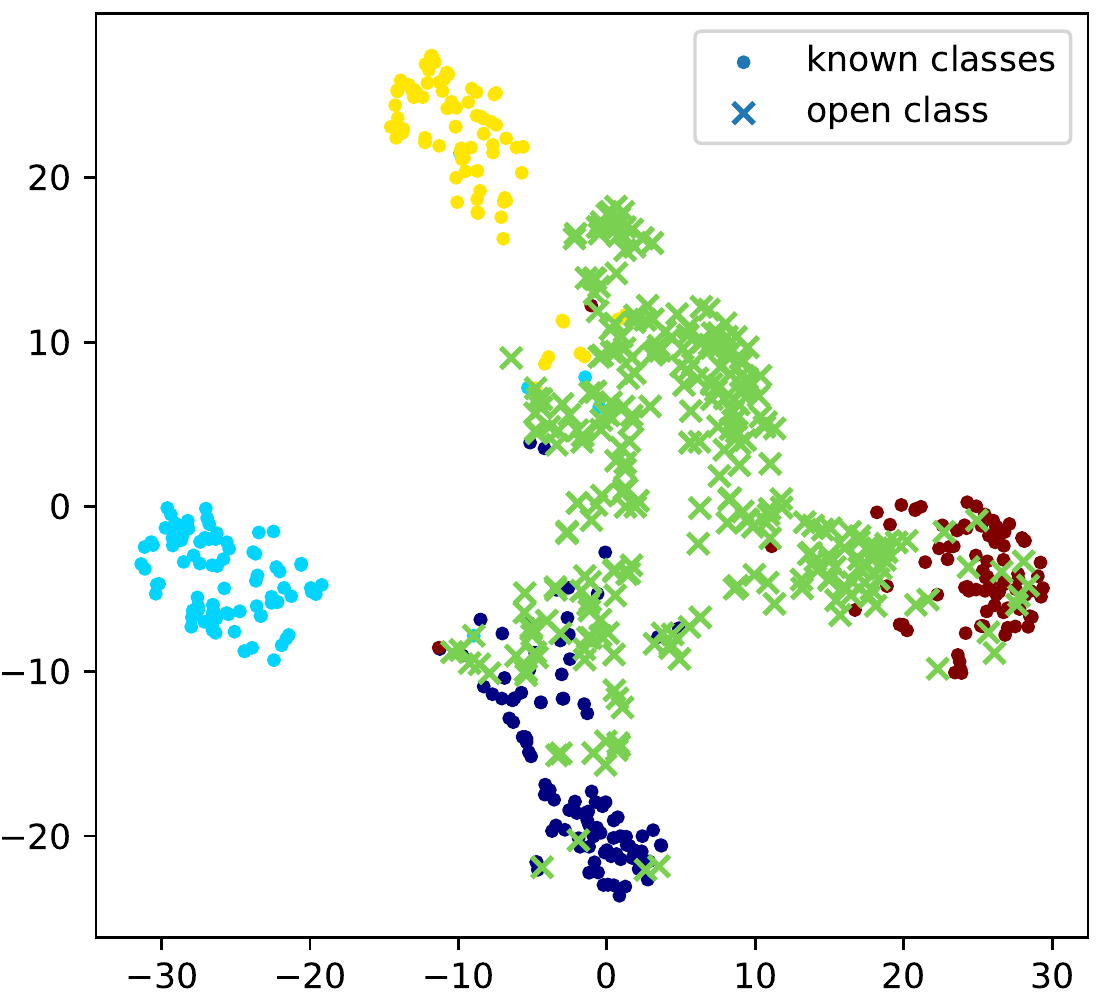}}\hspace{8mm}
  \subfigure[75\% known classes on CLINC.]{
        \includegraphics [width=0.27 \textwidth]{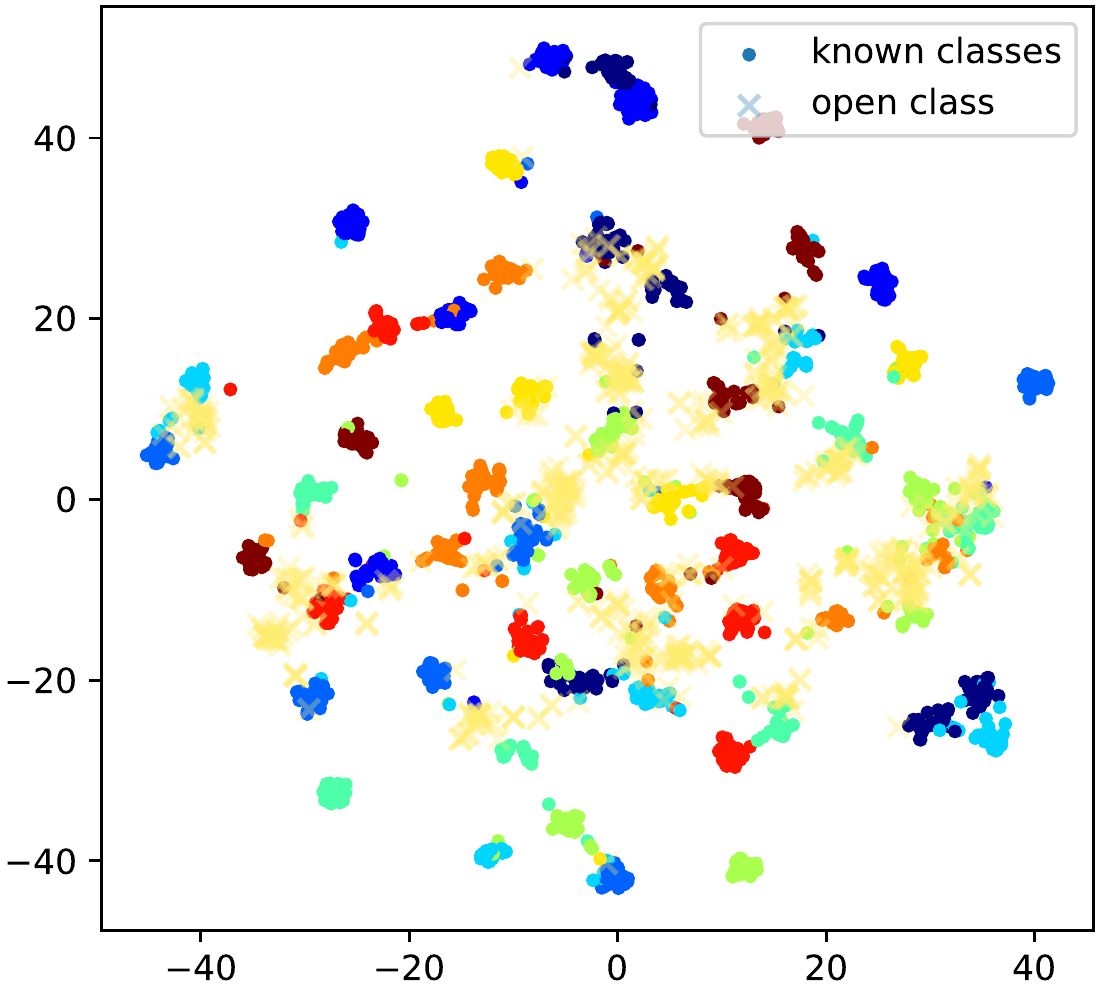}
        }\hspace{8mm}
        \subfigure[75\% known classes on OOS.]{
 \includegraphics [width=0.27 \textwidth]{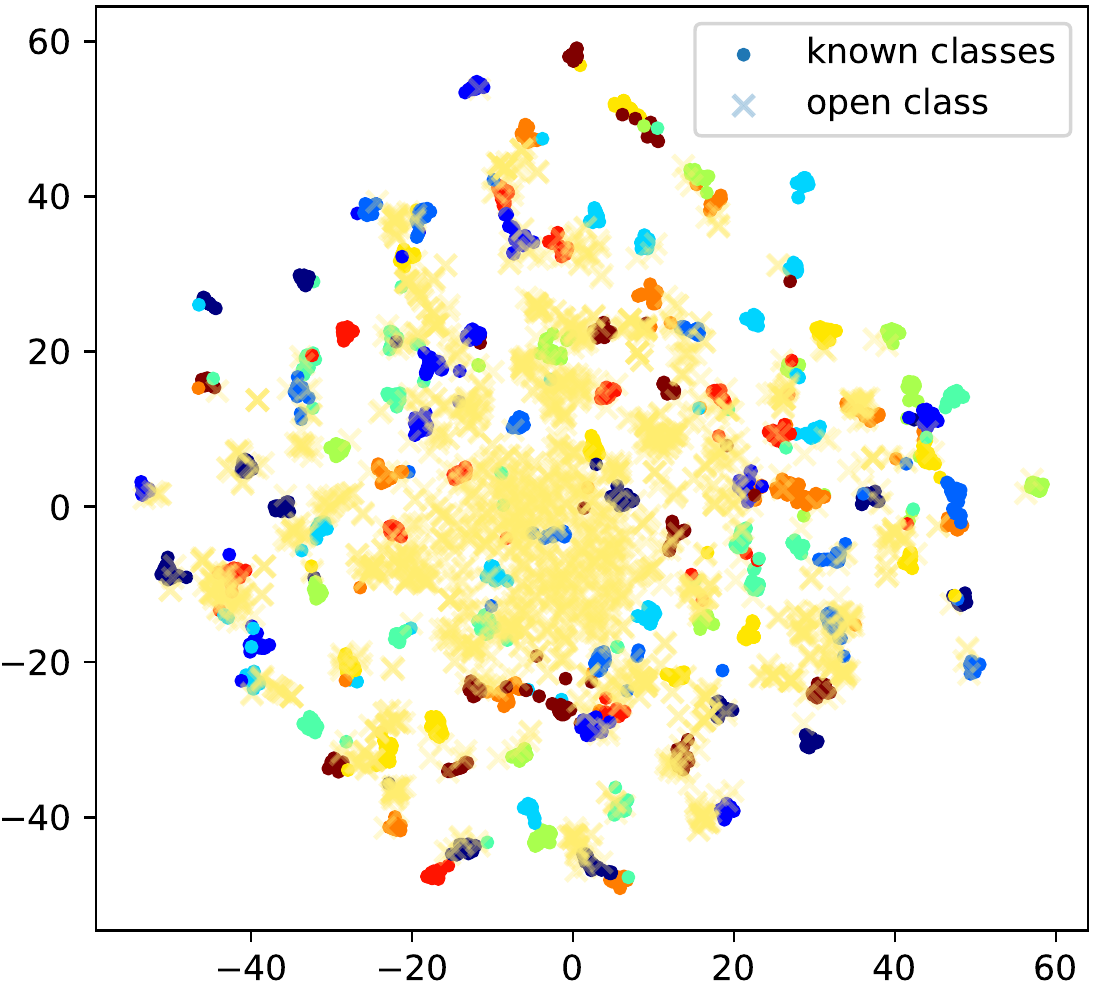}
  }
  \caption{Visualization of the learned embeddings from test set on three datasets.} \label{t-sne}
\end{figure*}

In this section, we perform error analysis on our model and report the confusion matrix under 50\% known classes on SNIPS dataset in Figure~\ref{cm}.
We can see that the off-diagonal elements can be divided into two types of errors: open class related error and known classes error.
First, we can find that open class related error occupies a large proportion of errors and many samples of open class have not been recognized.
For example, 49 samples of open class are predicted as label 0 and 74 samples of open class are predicted as label 3. 
There is also a small part of samples of known classes that are recognized as open class.
Second, there are five samples belonging to known classes error.
For example, 3 samples of class 0 are predicted as label 1 and 2 samples of class 1 are predicted as label 2.
This shows that our model can distinguish the known classes well, and the identification of the unknown classes is still the main challenge at present.
It is worth noting that it is hard to distinguish between label 3 and label 4.
By further observing the errors produced by the model on the SNIPS dataset, we found that some open intents are very similar with some known intents.
This makes it easy for these open intents to be classified as known intents, and thus producing errors.
For example, as shown in Table~\ref{case_study}, in the first case, it is hard to distinguish between the known intent `Play Music’ and the open intent `Add To Play list’.
This maybe because the utterances of these two intent classes are similar on vocabulary and semantics, and thus the features learned for these utterances are also hard to distinguish.
The second case shows a similar phenomenon.

\subsection{Visualization}

In this section, we visualize the embeddings of samples in test set by t-SNE \cite{van2008visualizing}.
As shown in Figure \ref{t-sne}, we plot three subfigures corresponding to three datasets: SNIPS, BANKING, and OOS, where the ratio of known classes are 50\%, 75\%, and 75\% respectively.
Among these subfigures, the samples of open class and known classes are represented as forks and dots, respectively.
From these subfigures, we can see that the learned embeddings can be separated well.
This shows that our method is effective.
In addition, we can also find that the samples of open class are mostly placed near known classes or between two different known classes.
This is consistent with our intuition of soft labeling and manifold mixup, and further shows the effective of our proposed two strategies.
Then this figure suggests, in agreement with results of Figure \ref{cm}, different known classes are better separated, and the main challenge is to identify open class.
It is worth noting that the difficulty of distinguishing between different known classes and open classes is different.

\section{Conclusion}
In this paper, we propose a deep open intent classification model based on soft labeling and manifold mixup.
Specifically, soft labeling gives each sample a probability of being predicted as an open intent, and manifold mixup generates open intent samples via interpolating between the hidden representation of two different known intent samples.
Through these two strategies, our model can directly perform (K+1)-class classification without outlier detection algorithms. 
We conduct extensive experiments on the benchmark datasets.
The experimental results demonstrate the effectiveness of our proposed method.

\ifCLASSOPTIONcaptionsoff
  \newpage
\fi



%
\bibliographystyle{IEEEtran}
\bibliography{TASLP.bib}

\end{document}